\documentclass{article}
\usepackage{microtype}
\usepackage{graphicx}
\usepackage{subfigure}
\usepackage{booktabs}

\usepackage[pagebackref,breaklinks,colorlinks]{hyperref}

\usepackage[accepted]{icml2023}

\usepackage{amsmath}
\usepackage{amssymb}
\usepackage{mathtools}
\usepackage{amsthm}
\usepackage{graphicx}
\usepackage{amsmath}
\usepackage{amssymb}
\usepackage{booktabs}
\usepackage{xcolor}
\usepackage{caption}
\usepackage{multicol,epigraph,wrapfig}

\usepackage[capitalize,noabbrev]{cleveref}

\theoremstyle{plain}
\newtheorem{theorem}{Theorem}[section]
\newtheorem{proposition}[theorem]{Proposition}

\theoremstyle{definition}

\theoremstyle{remark}

\usepackage[textsize=tiny]{todonotes}

\icmltitlerunning{Efficient RL via Disentangled Environment and Agent Representations}
\begin{document}

\twocolumn[
\icmltitle{Efficient RL via Disentangled Environment and Agent Representations}

\icmlsetsymbol{equal}{*}

\begin{icmlauthorlist}
\icmlauthor{Kevin Gmelin}{equal,CMU}
\icmlauthor{Shikhar Bahl}{equal,CMU}
\icmlauthor{Russell Mendonca}{CMU}
\icmlauthor{Deepak Pathak}{CMU}
\end{icmlauthorlist}

\icmlaffiliation{CMU}{Carnegie Mellon University}

\icmlcorrespondingauthor{Kevin Gmelin}{kgmelin11@gmail.com}
\icmlcorrespondingauthor{Shikhar Bahl}{sbahl2@andrew.cmu.edu}

\icmlkeywords{Machine Learning, ICML, Reinforcement Learning, Visual RL, Representation Learning for RL, Robotics}

\vskip 0.3in
]

\printAffiliationsAndNotice{\icmlEqualContribution}

\begin{abstract}
Agents that are aware of the separation between themselves and their environments can leverage this understanding to form effective representations of visual input. We propose an approach for learning such structured representations for RL algorithms, using \emph{visual knowledge of the agent}, such as its shape or mask, which is often inexpensive to obtain. This is incorporated into the RL objective using a simple auxiliary loss. We show that our method, \textbf{S}tructured \textbf{E}nvironment-\textbf{A}gent \textbf{R}epresentations (\ours), outperforms state-of-the-art model-free approaches over 18 different challenging visual simulation environments spanning 5 different robots. 
\end{abstract}

\section{Introduction}
Proprioception, i.e. the knowledge of one's own self, is heavily used by biological entities enabling them to perform various real-world tasks such as walking, manipulation, and navigation. Awareness of the position and movement of their body, and perceiving the environment as external to themselves, enables forming efficient representations of the observed input \cite{shapiro2010embodied}.
In contrast, most contemporary methods in visual reinforcement learning (RL) often learn combined representations in an end-to-end manner~\cite{kalashnikov2021mt, arulkumaran2017deep, kalashnikov2018qt, levineFDA15, peters2010reps} and require large amounts of data as a result.
Inspired by the concept of the interface between the ``inner" and ``outer" environments, we study the following question: is there a natural way to build a representation that can disentangle a robotic agent from its environment, and does that improve learning efficiency for RL? 

\begin{figure}[t!]
    \centering
    \includegraphics[width=\linewidth]{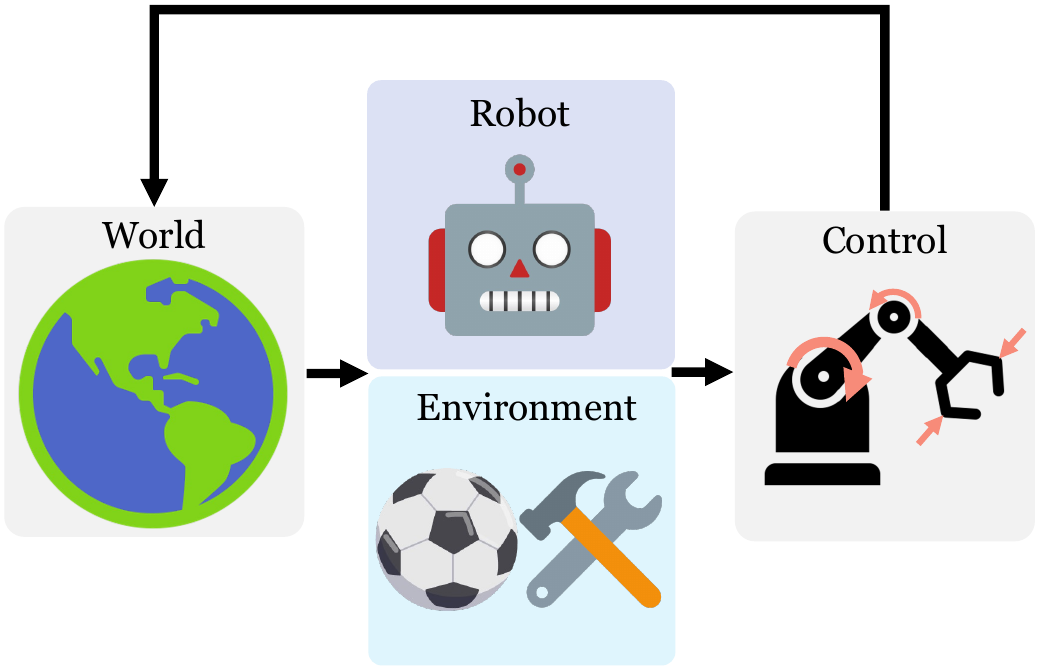}
    \caption{ \ours learns structured representations for visual control, by leveraging knowledge of the robot. } 
    \label{fig:teaser}
    \vspace{-0.1in}
\end{figure}

We argue that knowledge of the agent allows RL algorithms to focus on more interesting aspects of the visual input, such as objects in the environment. 
Thus, \textit{by explicitly forcing the learning algorithm to disentangle agent-centric representations, we also implicitly create environment-centric ones}. This allows for visual RL approaches to not only learn faster but makes the representations interpretable. Consider a policy trained on a door-opening task with one type of robot. If it is aware of the agent-environment distinction, then it will be able to adapt a lot faster when deployed on a new robot, or a new task like opening a drawer, since
it will not need to relearn visual appearance and it can focus just on dynamics. How can we incorporate this into an RL training setup? 

Our solution is simple: \textit{we augment the RL loss with an agent-centric auxiliary loss}. Naively doing so using proprioceptive data like joint angles or end-effector positions does not result in the desired representations, as these are not grounded in the visual observation space, and do not contain information about the robot's appearance. We find that the most grounded forms of supervision are agent \textit{masks}. If some information is known about the agent, it is often quite practical to obtain (even rough) masks for agents, for example, by training a segmentation model.

\begin{figure}[t!]
    \centering
    \includegraphics[width=\linewidth]{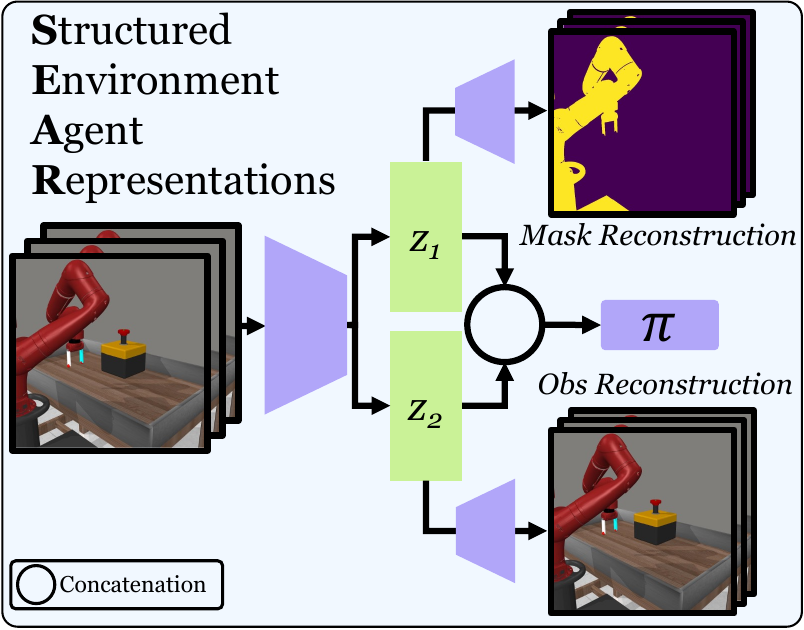}
    \caption{Overview of \ours. We use the mask of the agent to learn structured representations. }
    \label{fig:method}
    \vspace{-0.2in} 
\end{figure}

Our approach, \textbf{S}tructured \textbf{E}nvironment-\textbf{A}gent \textbf{R}epresentations (\ours), formulates the agent-environment representation learning problem as an auxiliary loss to the RL objective. This loss is the sum of the reconstruction loss of the agent mask and the full image as shown in Figure~\ref{fig:method}. We mathematically formulate the representation learning problem using variational inference and show that it leads to improved control.

Our approach is complementary to the prior works that aim to improve RL sample efficiency by adding auxiliary losses to the representations learned by the policy, drawing inspiration from advances in self-supervised learning in computer vision. This includes using inpainting losses~\cite{pathak2016context,he2022masked}, contrastive learning~\cite{he2020momentum, pmlr-v119-chen20j, chen2021exploring, grill2020bootstrap, pmlr-v119-laskin20a}, large-scale video pre-training \cite{r3m} or simple yet effective image augmentation \cite{yarats2021drqv2, laskin2020reinforcement} to make the network more robust. Such implicit inductive biases allow for useful representations, but they don't take advantage of the fact that there exists a lot of untapped knowledge of the \textit{agent}. Access to such knowledge is remarkably inexpensive as the agent morphology or joints are often known beforehand.

Empirically, we find that \ours outperforms state-of-the-art model-free approaches for RL on a large suite of 18 different challenging environments, spanning 5 different robots, including the sawyer, franka and adroit-hand robots. 

\section{Related Work}

\paragraph{RL from pixels} Model-based approaches \cite{hafner2019dream, hafner2020mastering, ebert2017self} learn an efficient latent space dynamics model to learn a policy for visual control tasks. Model-free algorithms such as RAD \cite{laskin2020reinforcement} and DrQ \cite{kostrikov2020image} make use of image augmentations to provide additional inductive biases. Some methods have focused on improving the representations learned for visual control through auxillary tasks, such as contrastive learning \cite{pmlr-v119-laskin20a}. Perhaps most similar to our algorithm, SAC-AE \cite{yarats2021improving} uses an autoencoder, in addition to losses from the critic in SAC \cite{haarnoja2018soft}, to train an image encoder used for visual control. However, unlike our algorithm, SAC-AE \cite{yarats2021improving} only uses a single decoder, and does not attempt to make any explicit distinction between agent and environment. DrQv2 \cite{yarats2021drqv2} is a state-of-the-art model-free reinforcement learning algorithm for visual continuous control, using random shift image augmentations and n-step returns for improved sample efficiency. 

\paragraph{Agent-Environment Centric Learning} Previous approaches have directly structured the representation space using forward or inverse dynamics \cite{zhang2019solar, watter2015embed}, but these methods do not scale well to challenging image-based manipulation tasks. \citet{hu2022know} learn a factorized visual dynamics model, using the analytical forward kinematics of the robot and a learned world model. While this enables transfer of the world model to new robots with a similar action space, it doesn't exploit agent knowledge while training.  Previous works have also implicitly trained policies aware of robot morphology, using the transferablility of the policies as signal \cite{yu2018one, duan2017one, dasari2021transformers, finn2017one, shaw2022video}. Some methods directly train separate robot and task modules, and attempt to transfer to new combinations of these modules \cite{neumann2014learning, devin2017learning}. It is also possible to construct a policy where each node is represented as a joint and each link is an edge \cite{wang2018nervenet, huang2020one}, allowing for efficient generalization to new morphologies. Decoupling environment and agent learning has also been popular in exploration-based approaches \cite{parisi2021interesting, hu2022know, bahl2022human2, bahl2023affordances, mendonca2023alan}.  Unlike work which learns self-perception, we assume access to a self-perception module to provide access to robot segmentation masks, and then investigate how such a model can be incorporated to improve policy learning. As we show in Appendix \ref{appendix:real_robot_masks}, such self-perception modules can be easily obtained.

\section{Background}

\paragraph{Reinforcement Learning} Formally, the RL problem is defined by a Markov decision process (MDP) $(\mathcal{S}, \mathcal{A}, \mathcal{T}, \mathcal{S}_0, \mathcal{R}, \gamma)$, where $\mathcal{S}$ is the state space, $ \mathcal{A}$ is the action space, $\mathcal{T}(s_{t+1}|s_t,a_t)$ is the unknown state transition function, $\mathcal{S}_0(s)$ is the initial state distribution, $\mathcal{R}(s_t, a_t)$ is the reward function, and $\gamma \in (0, 1)$ is the discount factor. An agent acts according to some policy $\pi(a|s)$ and the learning objective is to maximize the expected return, $\mathbb{E}_{s_t, a_t \sim \pi}\left[\sum_t \gamma^t \mathcal{R}(s_t, a_t)\right]$.

\paragraph{Representation Learning and RL} Learning effective control policies directly from raw image observations without instrumented setups to detect the states of different objects in the world is quite challenging. One approach to address this is to learn low dimensional representations for control \cite{zhu2020ingredients, nair2018visual, pong2019skew, laskin2020reinforcement}, bringing the setting closer to state-based control. These approaches often model the state at each timestep as a latent $z$, and learn an encoder $q_\theta$  and decoder $p_\phi$ by maximizing the evidence lower bound on the log likelihood of the image $X$: 

\vspace{-0.25in} 
\begin{equation}
    \log p(X) \geq \mathbb{E}_{z \sim q_\theta}{p_\phi(X | z)} - D_{KL}(q_\theta(z | X) || p(z))
\end{equation}

\section{Structured Latents for Control}

\subsection{Learning Structured Latents}

\begin{wrapfigure}{r}{0.5\linewidth}
    \vspace{-0.1in}
    \centering
    \includegraphics[width=\linewidth]{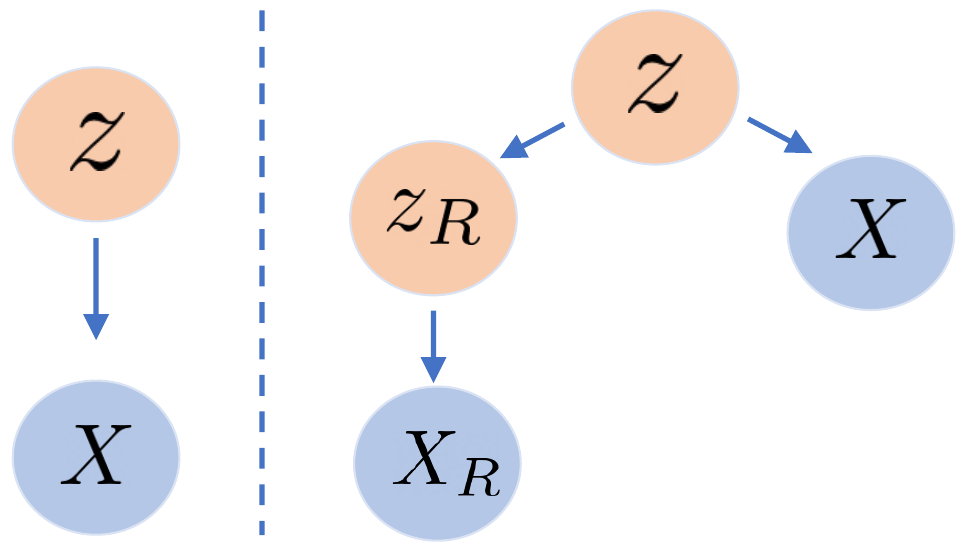}
    \caption{Graphical Models for prior approaches (left), vs \ours (right).} 
    \label{fig:pgm}
    \vspace{-0.1in}
\end{wrapfigure}

In the approaches described above, the learned latent $z$ models the entire image, consisting of both the robot and the objects in the environment. 
Is there a way we can learn more structured latent representations, given access to visual information pertaining to the robot $X_R$? In addition to learning variable $z$ which encodes the entire image, we also model $z_R$, which corresponds to salient information in $X_R$ for control. $z_R$ can be thought of as a processed version of $z$ which only contains \emph{agent}-relevant information, as opposed to information of the entire scene. To build the full dependency graph between variables $X, X_R, z, z_R$, we state the following desired property: 

\begin{proposition}
    \label{prop:pgm}
    If $z$ effectively encodes $X$, then $z_R$ and $X$ should be conditionally independent given $z$, i.e $p(z_R, X|z) = p(z_R|z).p(X|z)$
\end{proposition} 

We enforce this variable dependency by implementing the graphical model in Figure \ref{fig:pgm}. The joint probability distribution resulting from the model is :

\vspace{-0.2in}
\begin{equation}
    p(X,X_R, z,z_R) = p(z). p(z_R|z). p(X|z). p(X_R|z_R) 
\end{equation}

In order to learn $z,z_R$, we maximize $J = \log p(X, X_R)$, and learn a variational approximation $q_\theta(z, z_R|X)$ to the posterior, which leads to the following lower bound (derivation in Appendix \ref{appendix:rep_obj}): 

\begin{equation}
\begin{aligned}
\label{eq:elbo}
    \mathcal{L} & = \mathop{\mathbb{E}}_{z,z_R \sim q} \left[ \log p(X|z) \right] + \mathop{\mathbb{E}}_{z,z_R \sim q} \left[ \log p(X_R|z_R) \right] \\ 
                 & - D_{KL}(q(z,z_R|X))||(p(z,z_R)) \\
\end{aligned}
\end{equation}

Breaking each term down, we first want to maximize $\mathbb{E}_{z \sim q_\theta}{\log p(X | z)}$. This is the standard reconstruction loss of the Variational Autoencoder (VAE) \cite{kingma2013auto, rezende2014stochastic}. The second term, $\mathbb{E}_{z \sim q_\theta}{\log p(X_R | z_R)}$ reconstructs the provided robot visual information $X_R$ from  $z_R$, and the final term regularizes the posterior distribution just like in regular VAEs. 

\subsection{Effectiveness for Control}
Given these structured latent representations, do they enable better sample efficiency and faster training for policies? We propose the following thought experiment as a starting point. Under the graphical model described in Proposition \ref{prop:pgm}, consider a value function $V(z)$ that can, without loss of generality, be written as: 

\begin{equation}
    \label{eq:val}
    V(z) = V_R(z_R) +  V_C(z)
\end{equation}

$V_R$ is a function only dependent on the agent (the robot in our case) and $V_C$ is a coupling term for the value function, for the case where the agent is directly interacting with the environment. We argue that this formulation can represent a large class of value functions, especially under our disentangled representation space. 

Our main insight is that in contact rich tasks, there are many uncertainties in modeling the environment. Furthermore, the first step in many such tasks is control of the robot to move it to some target location before it can then engage in contact. Thus, for a large percentage of $t \in (1, T)$, there will be a large emphasis on the first term of Equation~\ref{eq:val}, which is the value function dependent on the agent directly. By explicitly modelling $z_R$, we expect that it will be easier for the agent to learn basic robot control. After the agent can perform basic control of the robot, and capture some reward only relevant to the motion of the robot, it can then move on to manipulating the environment, which is better captured with the second term of Equation~\ref{eq:val}. 

Overall, we hypothesize that modelling $z_R$ enables agents to be much more efficient in learning policies, which is supported by our experimental analysis. We provide more details and analysis on this in Appendix \ref{appendix:rl_performance}, and showcase this intuition in a toy experiment in Appendix \ref{appendix:toy_experiment}.

\subsection{Visually Grounded Agent Representations}

Now that we have seen how to learn agent-centric representations and that they are effective for learning policies, how do we obtain the agent-relevant observations $X_R$ required for representation learning? While proprioceptive data such as end-effector positions or joint angles are easily accessible, they are not visually grounded, and may not convey the full information needed to model how the robot appears in the image observations. Thus, we try to predict the visual robot observation in the image. One approach for this is to use a segmentation model of the agent to obtain a robot mask. This is reasonable to obtain for robots since the shape and appearance of the robot does not change across tasks, and obtaining the mask model is a one-time cost. Furthermore, they do not need to be fully accurate, and we show that our method can still learn effective control policies when using noisy or approximate masks. 

Given an image of the full scene $X$, we set $X_R$ to be a segmentation of the robot, $M$, where $M_{i, j} = 1$ for every pixel that is occupied by the agent and 0 otherwise. Using a visual encoder $q_\theta(.)$, we get $y = q_\theta(X)$, which is then split into 2 vectors to obtain $z_R$ and $z$. 

The encoder we use is similar to that of \citet{yarats2021drqv2}. $z_R$ is used as input to a decoder, $P_\phi(M|z_R)$ which tries to predict the robot mask. This decoder is trained using a binary cross-entropy loss, as shown in equation \ref{mask_loss}. Posing the agent-centric reconstruction as a classification problem allows for more efficient learning. 

\vspace{-0.2in} 
\begin{equation} \label{mask_loss}
    \mathcal{L}_{mask} = M \log{P_\phi(M|z_R)} + (1 - M) \log{(1 - P_\phi(M|z_R))}
\end{equation}

Following Equation~\ref{eq:elbo}, $z$ is used to approximate $p(X | z)$, by using a neural network decoder $P_\psi$, which tries to reconstruct the input image, $X$. This is trained using a mean squared error reconstruction loss, as shown in equation \ref{eq:reconstruction_loss}. 

\begin{equation} \label{eq:reconstruction_loss}
    \mathcal{L}_{recon} = \Vert X -P_\psi(z) \Vert _2 ^ 2
\end{equation}

We empirically found that setting the last term of Equation~\ref{eq:elbo} to be very low or 0 had much better results. 

While we use DrQv2 from \citet{yarats2021drqv2} as our base reinforcement learning algorithm, it is possible to apply \ours to any other reinforcement learning algorithm that utilizes a visual encoder, both for on or off-policy methods. We apply the same random shift to both the robot masks and input image observations as part of data augmentation. The mask decoder, reconstruction decoder, critic, and encoder are trained concurrently using a joint loss function, shown in Equation~\ref{eq:total_loss}, with coefficients weighing the relative importance of the three component losses. 

\begin{equation} \label{eq:total_loss}
    \mathcal{L} = \mathcal{L}_{critic} + c_1 \mathcal{L}_{recon} + c_2 \mathcal{L}_{mask}
\end{equation}

The overall architecture is shown in figure \ref{fig:method}. As with the DrQv2 algorithm, we do not let actor losses backpropagate into the encoder. Overall,  \ours introduces 4 additional hyperparameters to a reinforcement learning algorithm: two learning rates corresponding to two decoders, and two coefficients for the reconstruction and mask losses. To simplify hyperparameter tuning, we used the same learning rate as the critic for the two decoders.  Our approach can be seen in Algorithm~\ref{alg:method}. 

\begin{algorithm}[t!]
    \caption{SEAR: Structured Environment and Agent Representations for Control}
    \label{alg:method}
    \begin{algorithmic}[1]
        \FOR{t = 1 \ldots T} 
        \STATE Collect transition $(x_t, m_t, a_t, R(x_t, a_t), x_{t+n})$
        \STATE $\mathcal{D} \leftarrow \mathcal{D} \cup (x_t, m_t, a_t, R(x_t, a_t), x_{t+1})$
        \STATE \textsc{UpdateCriticAndDecoders}($\mathcal{D}$)
        \STATE \textsc{UpdateActor} \cite{yarats2021drqv2}
        \ENDFOR
        
        \FUNCTION{\textsc{UpdateCriticAndDecoders}($\mathcal{D}$)}
        \STATE $(x_t, M_t, a_t, r_{t:t+n-1}, x_{t+n}) \sim \mathcal{D}$
        \STATE Sample augmentation $A_1$
        \STATE $z_t\leftarrow q_{\theta}(A_1(x_t))$
        \STATE $[z_t^1; z_t^2] \leftarrow z_t$ 
        \STATE $\mathcal{L}_{mask} \leftarrow \mathcal{L}_{\text{ent}}(P_\phi(z_t^1), A_1(M_t))$
        \STATE $\mathcal{L}_{recon} \leftarrow ||P_\psi(z_t^2) - A_1(x_t))||_2^2$
        \STATE Compute $\mathcal{L}_{critic}$ \cite{yarats2021drqv2}
        \STATE $\mathcal{L}_{total} \leftarrow \mathcal{L}_{critic} + c_1 \mathcal{L}_{recon} + c_2 \mathcal{L}_{mask} $
        \STATE Update $\theta_{enc}, \theta_{critic}, \theta_{mask}, \theta_{recon}$ using $\mathcal{L}_{total}$ 
        \ENDFUNCTION
    \end{algorithmic}
\end{algorithm}

\section{Experiments}

In our experiments we seek to test the effectiveness of our structured representations in the following settings : 1) RL for manipulation with robot arms 2) RL for dexterous manipulation 3) Control in visually distracting environments 4) Transfer learning/finetuning. We also demonstrate some preliminary results in 5) multi-task learning. In this section, we describe our experimental setup, evaluations, baselines and results. 

\begin{figure}[h!]
    \centering
    \includegraphics[width=0.9\linewidth]{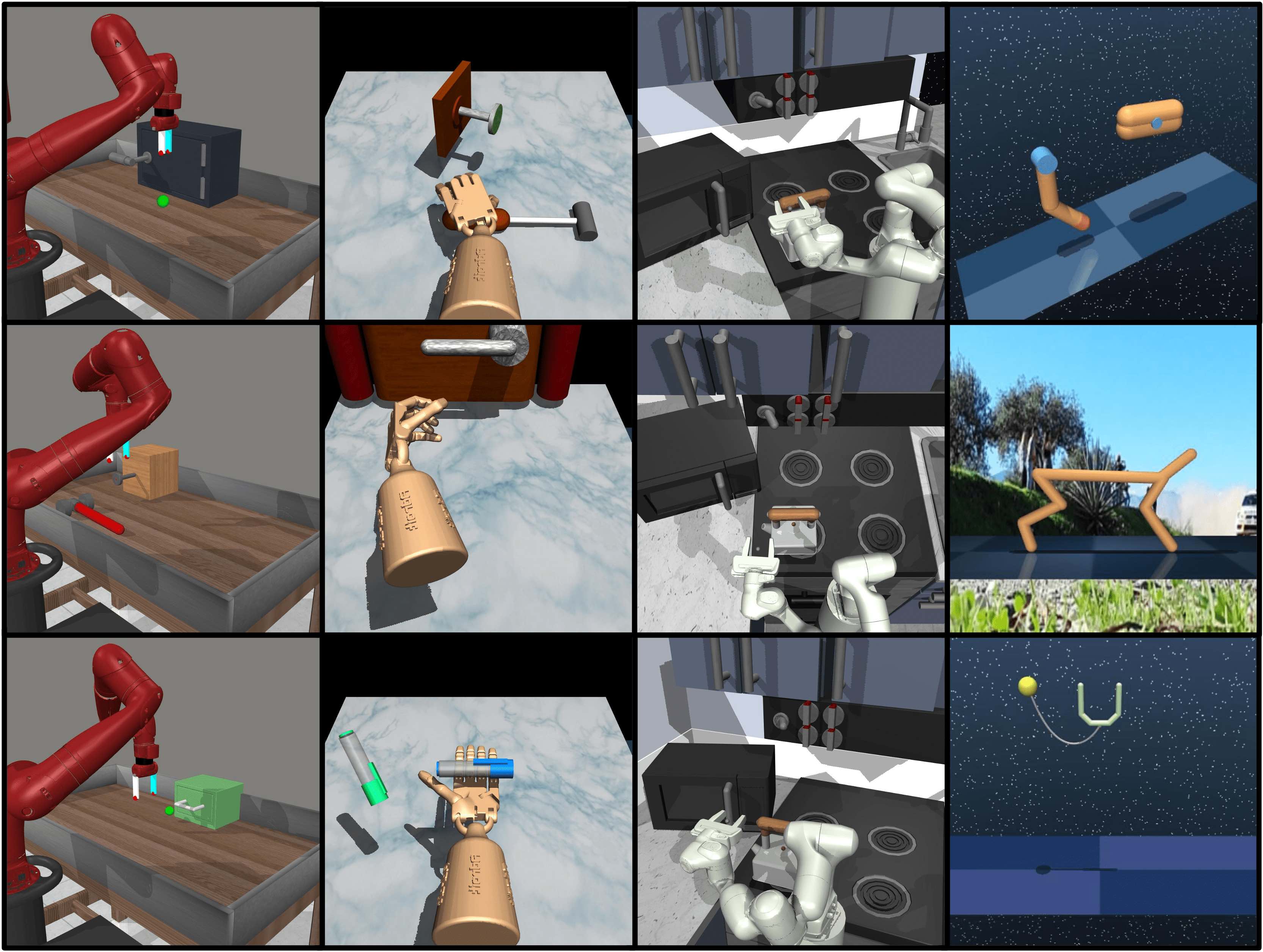}
    \caption{Environments used in the evaluation of SEAR. From left to right: Meta-World, Hand Manipulation Suite, Franka Kitchen, Distracting Control Suite} 
    \label{fig:tasks_3x4.png}
    \vspace{-0.1in}
\end{figure}

\paragraph{Environments:} We use the following Mujoco-simulated \cite{todorov12mujoco} environment suites:
\begin{itemize}
  \item \texttt{Meta-World} \cite{yu2020meta} - Table-top manipulation tasks performed by a Sawyer robot arm. 
  \item \texttt{Franka Kitchen} \cite{gupta2019relay} - Manipulating objects in a realistic kitchen with a Franka arm.
  \item \texttt{Hand Manipulation Suite} \cite{rajeswaran2017learning} - Manipulating objects with an Adroit hand.
  \item \texttt{Distracting Control Suite} \cite{stone2021distracting} - A variant of the DM Control suite \cite{tassa2018deepmind} with distractions added.
\end{itemize}

Note that we used implementations of the Hand Manipulation Suite and Franka Kitchen from the D4RL benchmark \cite{fu2020d4rl}. Images of these four environments are shown in figure \ref{fig:tasks_3x4.png}.  

\begin{figure}[t]
    \centering
    \includegraphics[width=\linewidth]{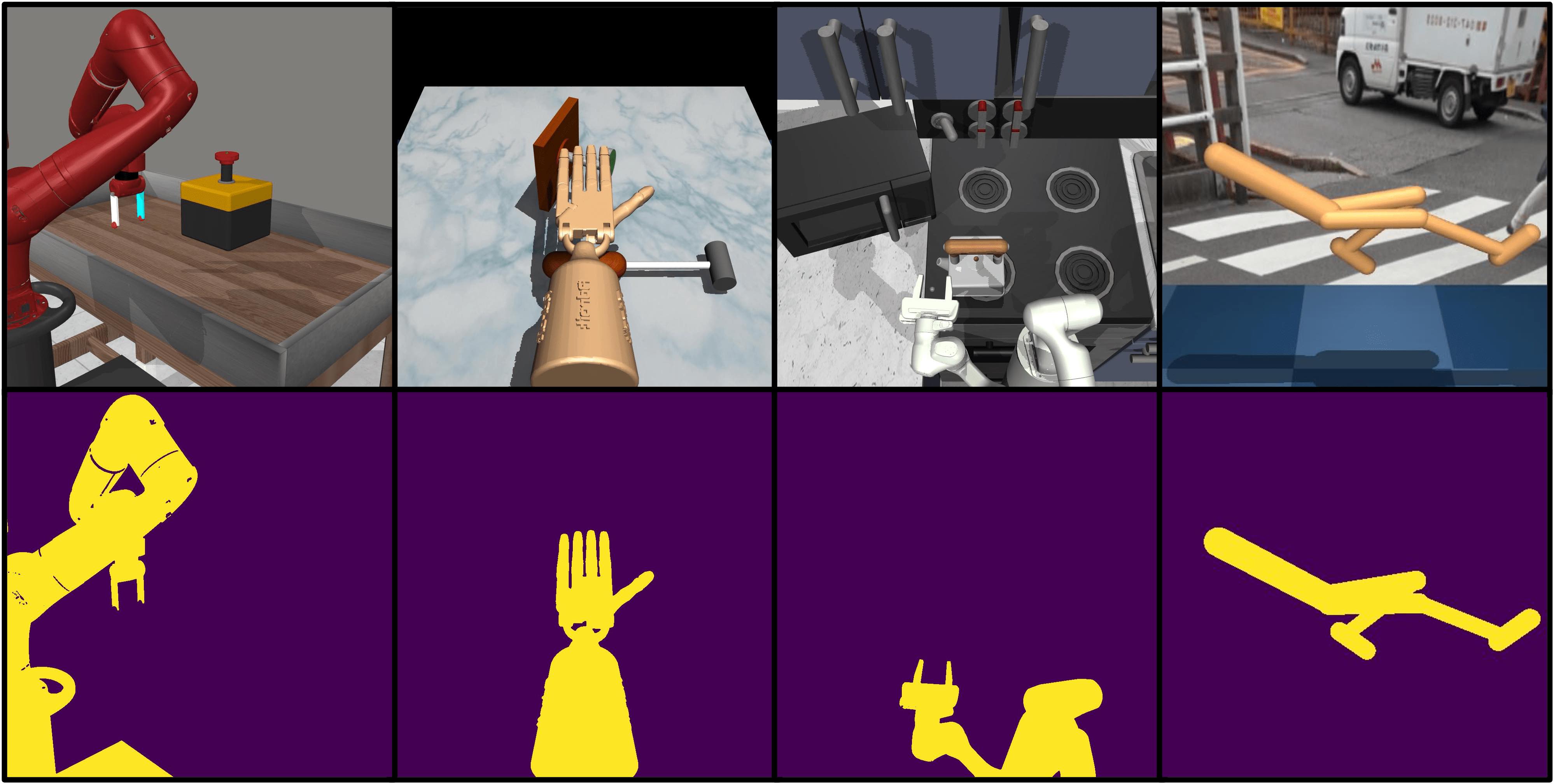}
    \caption{Environment RGB images with their corresponding robot segmentation masks.} 
    \label{fig:tasks_with_segmentation}
    \vspace{-0.1in}
\end{figure}

\paragraph{Setup:} We obtain agent masks from the simulator directly. Ground truth segmentation masks were generated using Mujoco's rendering API, which has a flag for rendering segmented images. Pixel values from these segmentation renders correspond to geometric IDs. We then create binary masks based on geometric IDs known to correspond to the robot for a given environment. Example masks from each suite are shown in Fig. \ref{fig:tasks_with_segmentation}. Note that one could also use a segmentation model for these. The agent receives an RGB image as input, and a mask as supervision for the representation learning (all of which are 84x84). For each environment, a frame stack of 3 was used. For Meta-World and Distracting Control tasks, an action repeat of 2 was used, whereas an action repeat of 1 was used for Hand Manipulation Suite and Franka Kitchen tasks. In the Franka Kitchen setup, we used a sparse reward of 1 for each of the individual tasks. For the multi-task version, we add the sparse reward from each subtask achieved. 

We compare to the following baselines in our experiments - 
\begin{itemize}
    \item \texttt{DrQ} - DDPG \cite{lillicrap2015continuous} with image augmentations, shown to be state of the art in many visual RL settings. We use the code and setup from \citet{yarats2021drqv2}, and implement \ours on top of this base. 
    \item \texttt{DrQ-AE} - Uses the same encoder as \ours, but without the mask decoder. Has a single decoder operating on the entire latent vector in order to reconstruct the original input image. This is a version of SAC-AE \cite{yarats2021improving}. We run this to test 
    if agent-environment disentangling is helping or if any form of reconstruction will have the same effect.
    \item \texttt{DrQ-MAE} - Reconstructing randomly masked patches in the images. Such methods have shown to be robust in self-supervised learning and control settings \cite{mvp, radosavovic2022real, he2022masked}.
    \item \texttt{CURL} - State-of-the-art self-supervised visual learning for RL approach \cite{pmlr-v119-laskin20a}, which has been shown to be a useful auxiliary loss. Performs momentum contrastive learning \cite{he2020momentum} on the input images during the RL loop. 
\end{itemize}

\subsection{Analysis and Ablations}
\label{subsection:AnalysisAndAblations}

\begin{figure}[h!]
    \centering
    \includegraphics[width=\linewidth]{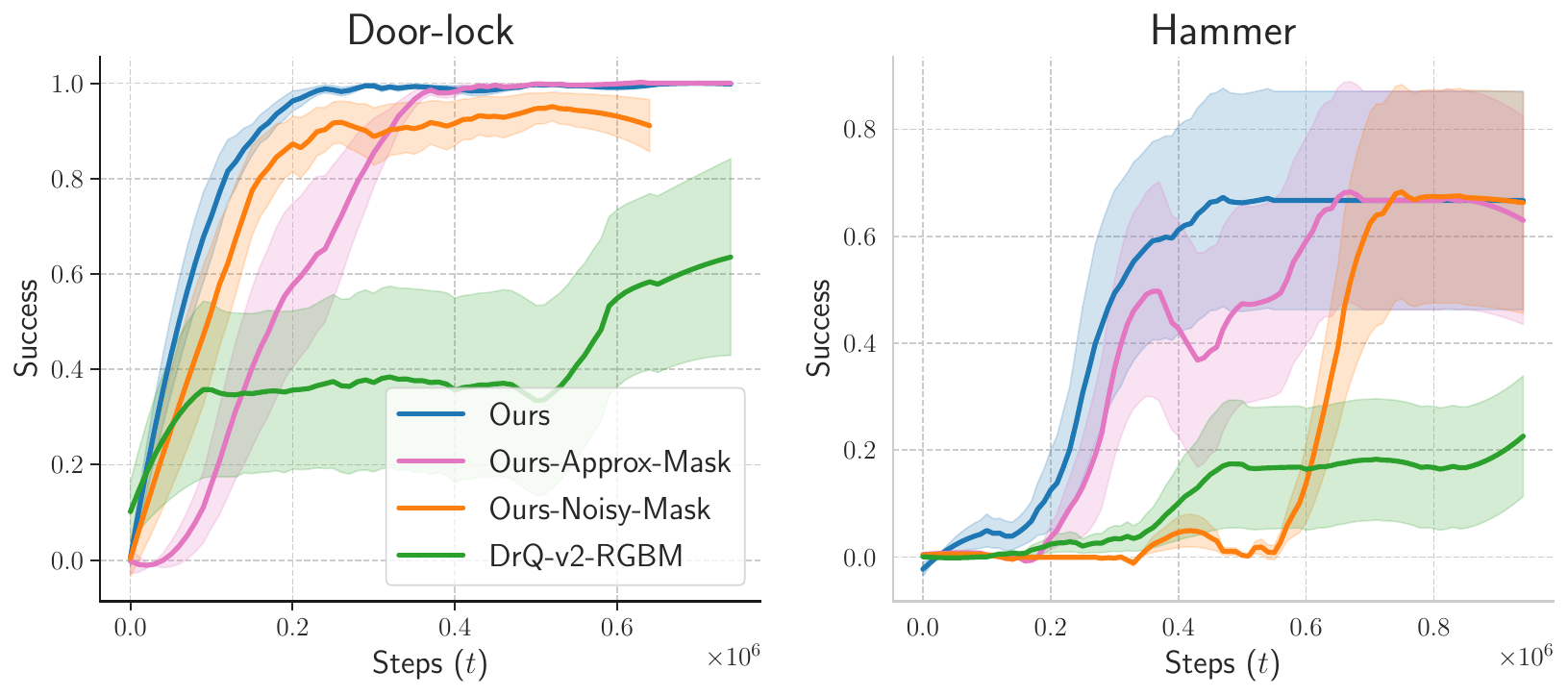}
    \caption{Analysis of robustness of \ours with noisy and approximate robot masks.}
    \label{fig:noise}
    \vspace{-0.1in}
\end{figure}

\begin{figure}[h!]
    \centering
    \includegraphics[width=0.9\linewidth]{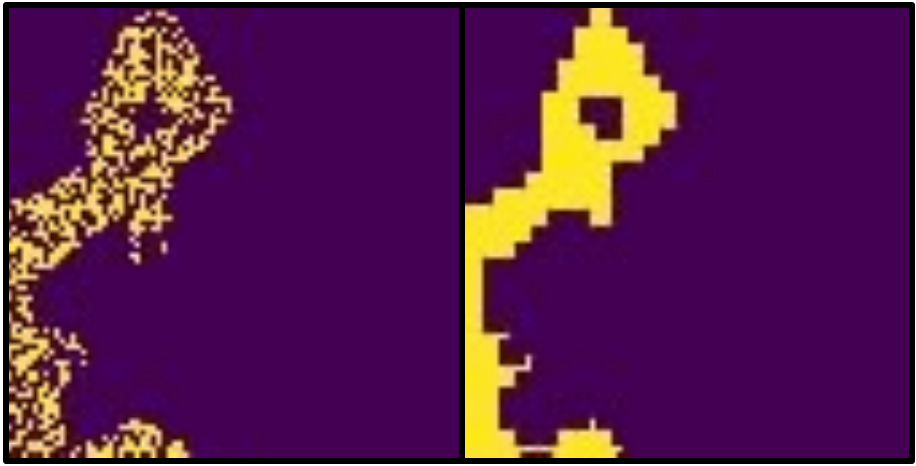}
    \caption{Noisy Mask (left) and Approximate Mask (right). Implementation described in Appendix \ref{appendix:noisy_mask_generation}}
    \label{fig:masks-noise}
    \vspace{-0.1in}
\end{figure}

\begin{figure*}[h!]
    \centering
    \includegraphics[width=\linewidth]{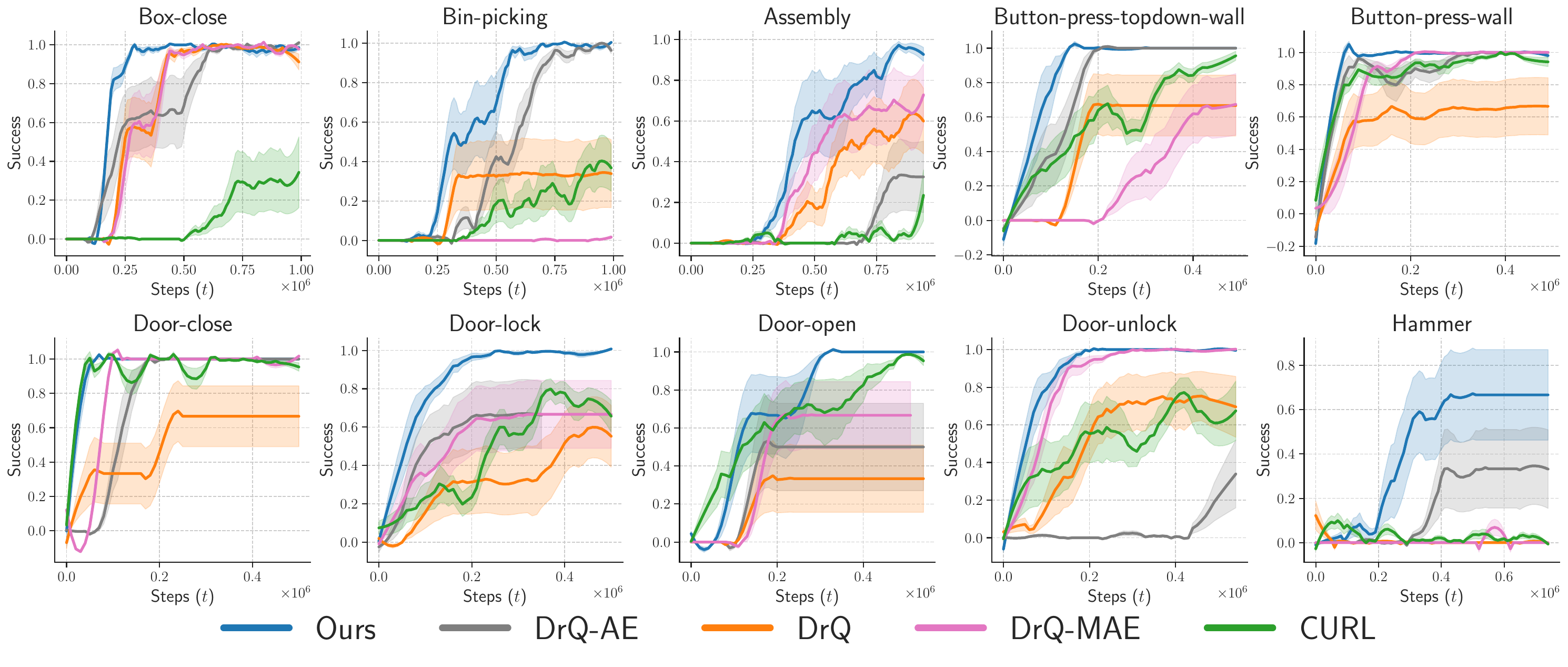}
    \caption{\textbf{Meta-World}: Success rate for different methods on 10 different tasks from the Meta-World suite. \ours outperforms or matches prior approaches on all tasks.}
    \label{fig:mtw}
    \vspace{-0.1in}
\end{figure*}

\paragraph{Robustness to inaccuracies in robot masks:} In real world scenarios, we do not have direct access to robot segmentation masks and may need to either have a trained model or use robot proprioception data. While \ours does use additional information, which is the robot mask data, we argue that this information is not difficult to obtain or approximate. To test this, we run multiple ablations of our model and analyze these. Firstly, we confirm that adding even an approximate amount of agent information helps in the case of learning. This is done via two experiments shown in Figure~\ref{fig:noise}. Agents were trained on two separate Meta-World tasks. We artificially introduce noise into the masks to evaluate the robustness of our approach (shown in orange). Another option is to use proprioceptive data to get an approximate mask. We mimic this by generating a large patch of pixels around the joints to get some mask that has a somewhat similar shape to the robot. This curve is shown in magenta in Figure~\ref{fig:noise}. With both, we see that the performance does not decrease by much, almost being similar to \ours with a perfect mask, even when a large amount of noise is added. Such masks can be seen in Figure~\ref{fig:masks-noise}. This experiment allows us to conclude that when training in the real world, a segmentation model of the robot can be used, even if it has inaccuracies. In Appendix \ref{appendix:real_robot_masks}, we further qualitatively show that a simple masking model fine-tuned on 100 images (including those collected on our robots, and internet images) can give good robot masks for many out-of-distribution images. 

\begin{figure*}[h!]
    \centering
    \includegraphics[width=\linewidth]{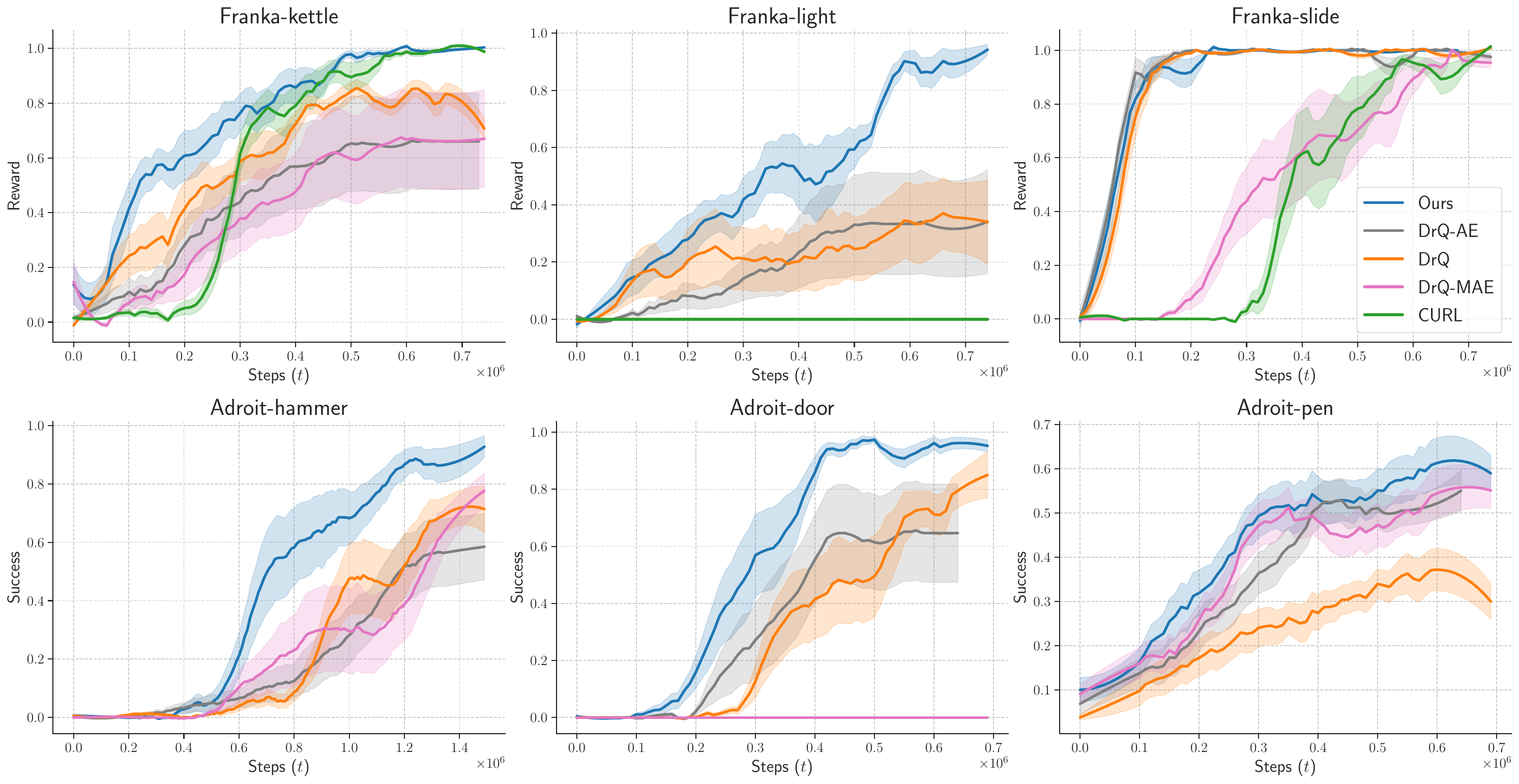}
    \caption{\textbf{Franka Kitchen \& Adroit}: Success for different methods on 3 different tasks from the FrankaKitchen suite (top) and from the Adroit-hand suite (bottom). \ours outperforms or matches prior approaches on all tasks.}
    \label{fig:frnk}
    \vspace{-0.1in}
\end{figure*}

\paragraph{Implicit Representations:} Concluding that incorporating the robot information is important, we analyze other ways to learn $z_R$ from the robot mask, for example in an implicit manner. Thus, we run a simple ablation (green) with \texttt{DrQ-RGBM}. This approach uses the DrQ-v2 algorithm trained on observations consisting of RGB images concatenated with robot masks. We see that this performance on both environments in Figure~\ref{fig:noise} is worse than that of our approach. We also see that \ours is robust to hyperparameters such as $c_1$ and $c_2$, the loss coefficients for the reconstruction and mask (Appendix \ref{appendix:decoder_ablations}). In Appendix \ref{appendix:no_zr}, we further show that there is a performance drop if the mask loss is added as an auxillary loss without splitting the latent vector.

\subsection{Continuous Control Experiments}

For the continuous control experiments, we evaluate \ours on various challenging environments, tasks and benchmarks, where the agent varies greatly in morphology, ranging from a 7-DoF robot arm to a hand or an embodied two dimensional walker, all in MuJoCo \cite{todorov12mujoco}. We firstly evaluated our agent on 10 different visually challenging tasks from the Meta-World benchmark \cite{yu2020meta}, as well as on three hand manipulation suite tasks, three Franka Kitchen tasks, and two Distracting Control tasks.  We measure and report either episode success or total reward, at test time. 

\paragraph{Meta-World} We compared our agents performance, measured as episode success rate, to baselines on single-task manipulation tasks available in the Meta-World benchmark. Due to compute and time constraints, we only ran experiments on 10 tasks, which we chose because they seemed like the most contact-rich. The results are shown in figure \ref{fig:mtw}. We observed greater sample efficiency on many of the tasks. These are all visually diverse and relatively challenging settings. Furthermore, on three of the tasks, \ours achieved the highest episode reward by the end of training. On button-press-wall and door-close, \ours did not outperform all the baselines, but it still matched the best ones. We found the auto-encoding baseline \texttt{DrQ-AE} to be a strong one, while contrastive learning (\texttt{CURL}) had difficulty. How does \ours scale to different types of robots, for example, a Franka? 

\paragraph{Franka Kitchen} We evaluated our agent on the Franka Kitchen benchmark, a diverse and visually challenging environment, where a task is defined by a set of objects in the scene that have to be moved to pre-specified goal locations. We looked at three different, single item-tasks: kettle, light, and slider. A reward of 1 is given for successfully moving a task item to its pre-specified goal location. Results, given as total episode reward, are shown in figure \ref{fig:frnk}. \ours beat all baselines on light, and matched the best baseline on the other two tasks. Note that the best baseline was different between these two tasks (kettle and slide), suggesting that while there is high inter-task variance in the performance of the various algorithms, \ours tends to match or surpass the performance of the best baseline for a given task.  

\paragraph{Adroit Hand Manipulation}We evaluated episode success rate of our agent on three different tasks within the Adroit Hand Manipulation Suite: Pen, Hammer, and Door. Results are shown in figure \ref{fig:frnk}. \ours outperformed baselines on two of the tasks, and matched the best baseline on the other task. We also ran experiments on the Relocate task, but all of the agents failed to learn, and thus we have not included it. We found that \texttt{CURL} was unable to solve any of the Adroit Hand Manipulation tasks, even with $>$ 1.5M environment steps. 

\begin{figure}[h!]
    \centering
    \includegraphics[width=\linewidth]{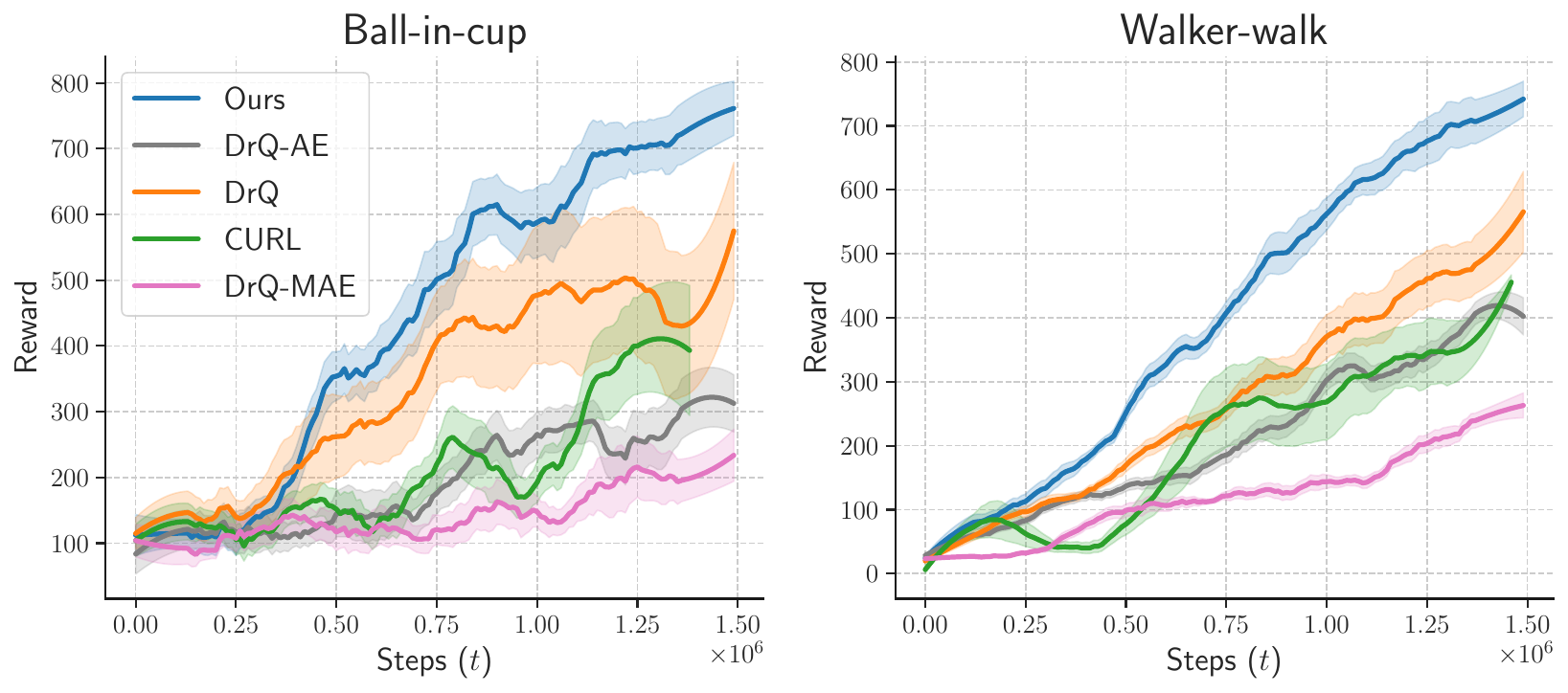}
    \caption{\textbf{Distracting Control:} Reward for different methods, \ours outperforms all the baselines.}
    \label{fig:distract}
\end{figure}

\paragraph{Distracting Control} If our hypothesis of \ours learning environment-agent disentangling representations was correct, we would see a strong performance in cases where control completely disentangles environment from agent. Thus we conducted an experiment on two challenging environments from the Distracting Control environments: \cite{stone2021distracting} the ball-in-cup catch and walker-walk, where each episode, a random image is shown in the background. Total episode reward results are shown in Fig \ref{fig:distract}. By the end of training, \ours achieved a higher total episode reward on both tasks compared to all other baselines. This suggests that \ours learns useful disentangled representations when controlling a robot in visually distracting environments.

\begin{figure*}[h!]
    \centering
    \includegraphics[width=\linewidth]{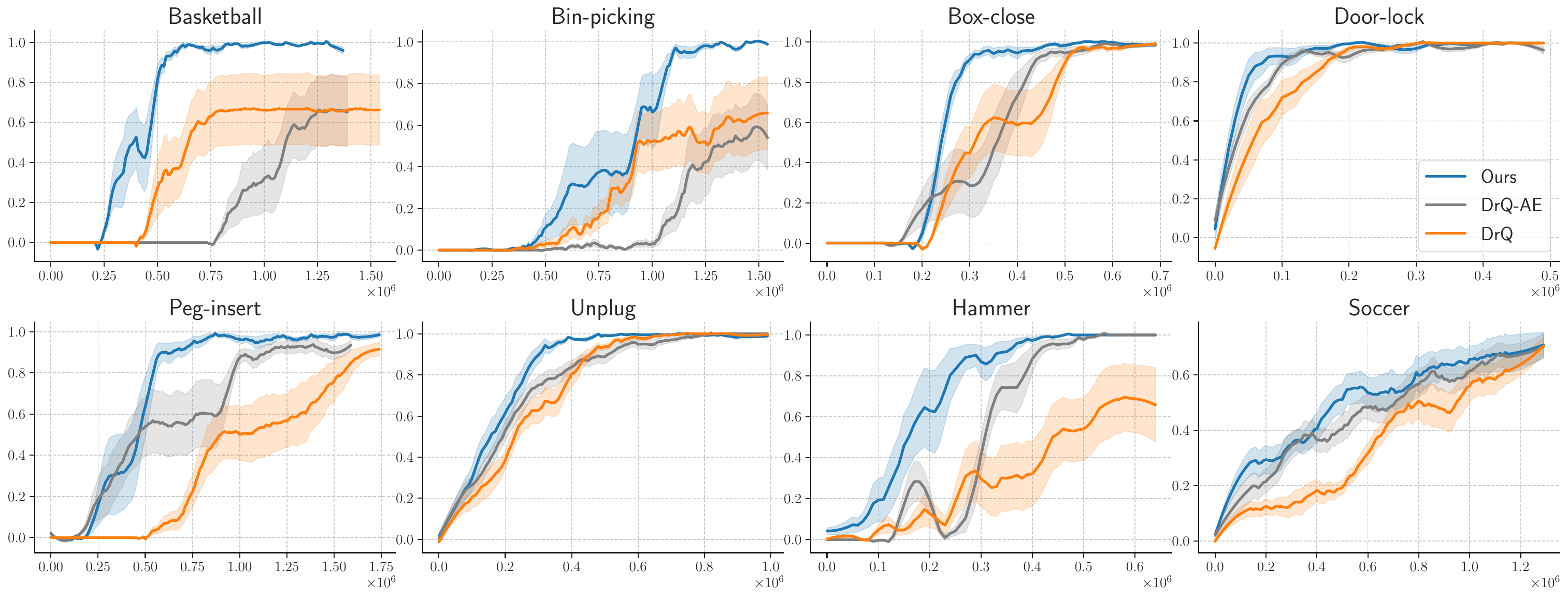}
    \caption{\textbf{Transfer}: Success for finetuning a multi-task policy pretrained on the Meta-World MT-10 suite to 8 different unseen tasks. \ours transfers representations much better than prior methods.}
    \label{fig:transfer}
    \vspace{-0.1in}
\end{figure*}

\paragraph{Transfer Learning} In transfer learning environments, while the task may be different, the agent is often the same. This leads us to expect that making an agent-environment distinction in representation learning will lead to improved performance during transfer learning. To test this, we pre-trained agents on a multi-task setting (MT10 suite), and then fine-tuned on new Meta-World tasks not part of the original MT10 tasks. We also fine-tuned on Peg-Insert, which the agents failed to learn in the multi-task setting (Potentially affected by a poor camera angle). Results for transfer learning are shown in figure \ref{fig:transfer}. \ours outperformed the baselines on many of the tasks, and matched the best baseline on the rest.  Overall, the results suggest that \ours learns a representation that performs well in transfer learning settings. We leave it to future work to investigate to what extent transfer learning performance is attributable to better transferability vs the better performance that \ours has shown in single-task learning.
 
\begin{figure}[h!]
    \centering
    \includegraphics[width=\linewidth]{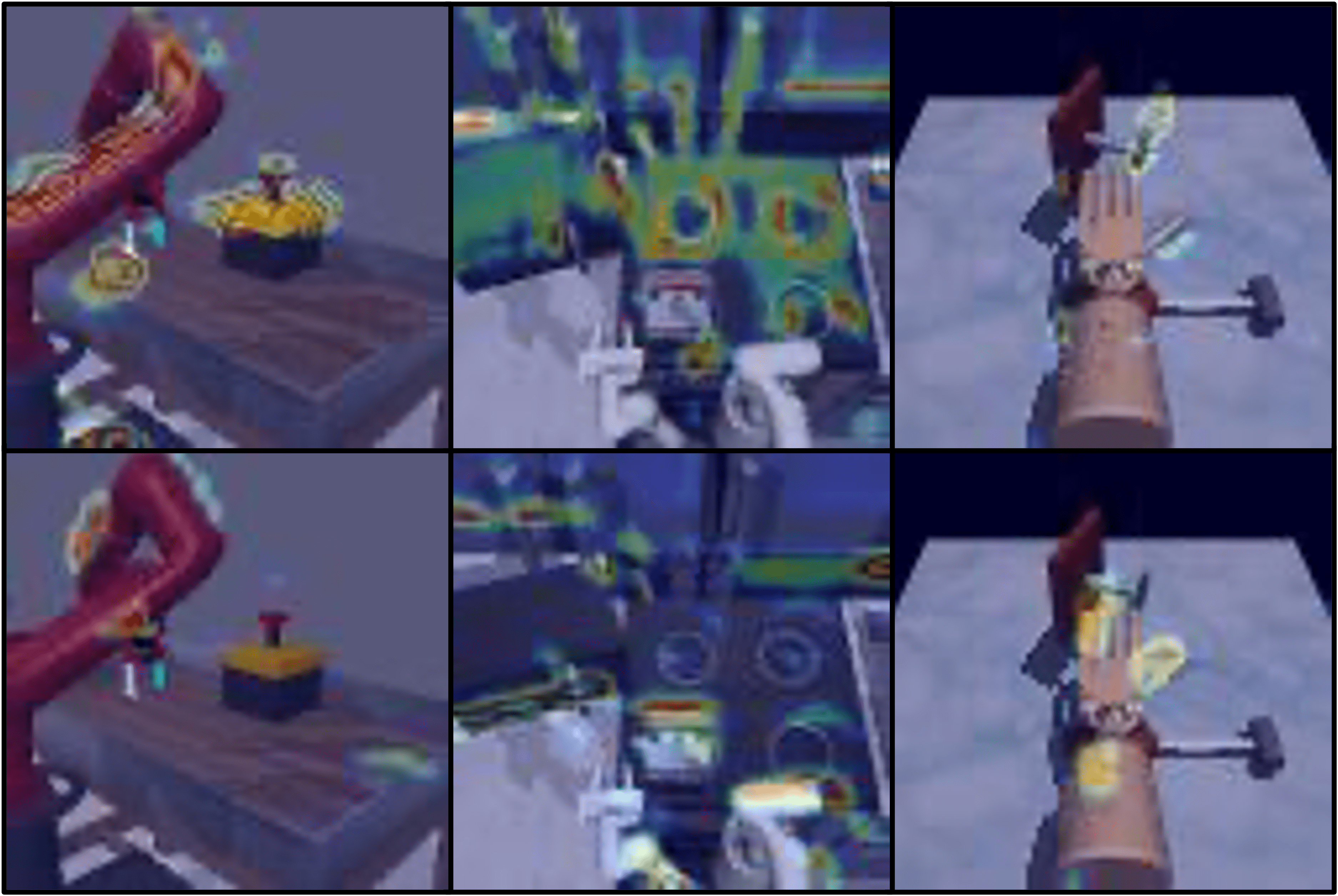}
    \caption{\ours activation maps for select channels from the last convolution layer in the encoder of \ours.}
    \label{fig:activations}
\end{figure}

\paragraph{Visualizing Feature Maps} To gain more insight into the difference in representations learned by \ours, we examine the activation maps of the encoder on one task each from three of our robotics environments. This activation map is the ReLU activations after the last encoder convolutional layer. Within the encoder, such activation maps would pass though an additional pooling and projection layer to form the latent vector. These activation maps are resized to the input image size, normalized to [0,1], and then blended with the original image after being mapped through a color map. We examine these activations since it is easier to visualize their spatial correlation with the input image, compared to visualizing the latent vector. Figure \ref{fig:activations} contains a few example maps, where we can see that \ours has more activity in regions around the robot in some maps and a lot more activity around objects in other maps. This is particularly salient in the Franka-Kitchen and Meta-World environments. In Appendix \ref{appendix:activation_maps}, we present a comparison of all 32 channels for both \ours and \texttt{DrQ}. While they don’t show full disentanglement (since the linear projection can map them in arbitrary combinations), we found it interesting that activations from the SEAR encoder tend to have more filters that fire on either the robot or the environment (Figures \ref{fig:mt_button_act_full} and \ref{fig:adroit_act_full}), or focus on more diverse parts of the environment (kettle, burners, cabinet or microwave in Figure \ref{fig:kitchen_act_full}) compared to DrQ filters which only fire on the kettle. We hope to study this type of disentanglement in more detail in future work. 

\begin{figure}[h!]
    \centering
    \includegraphics[width=\linewidth]{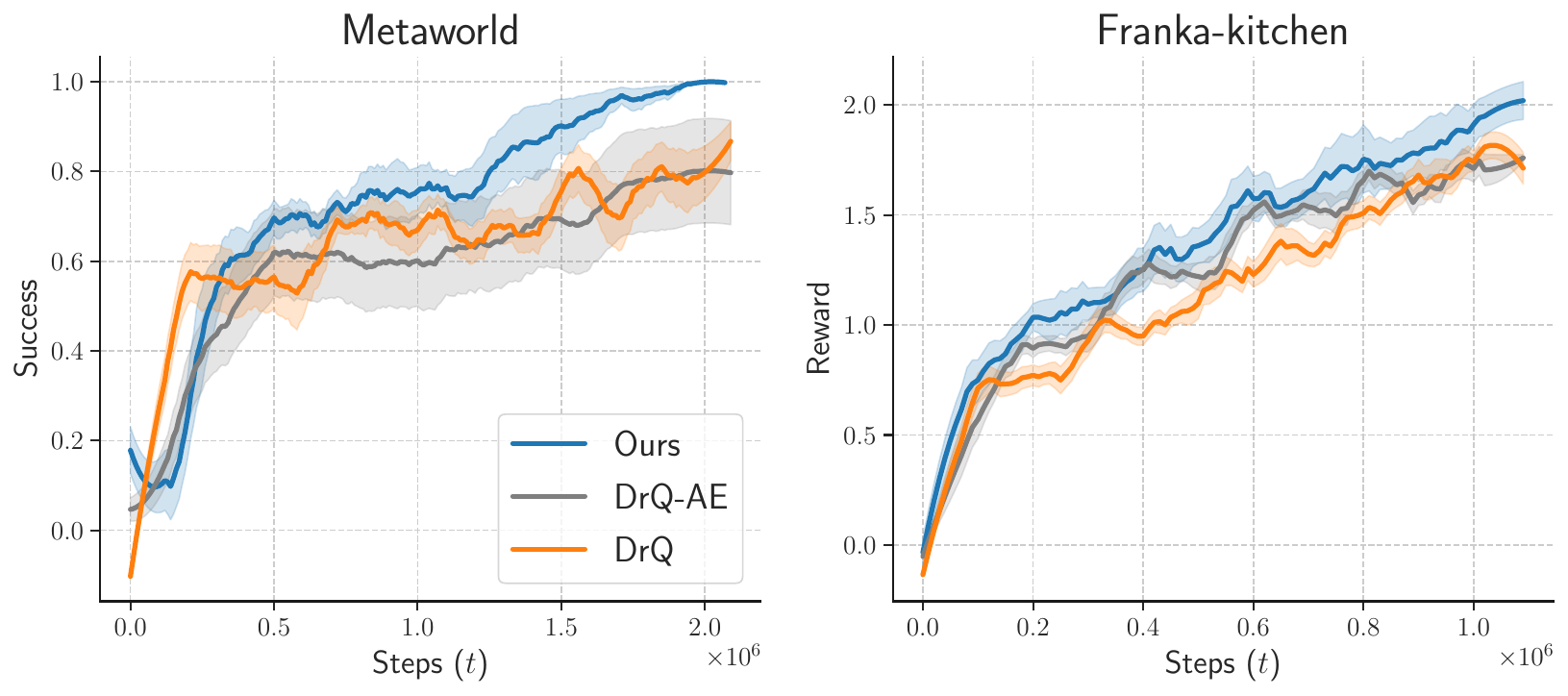}
    \caption{\textbf{Small-Scale Multi-task settings:} Success for different approaches for joint training on multiple tasks from an environment suite. }
    \label{fig:multi-task}
    \vspace{-0.1in}
\end{figure}

\begin{figure}[h!]
    \centering
    \includegraphics[width=\linewidth]{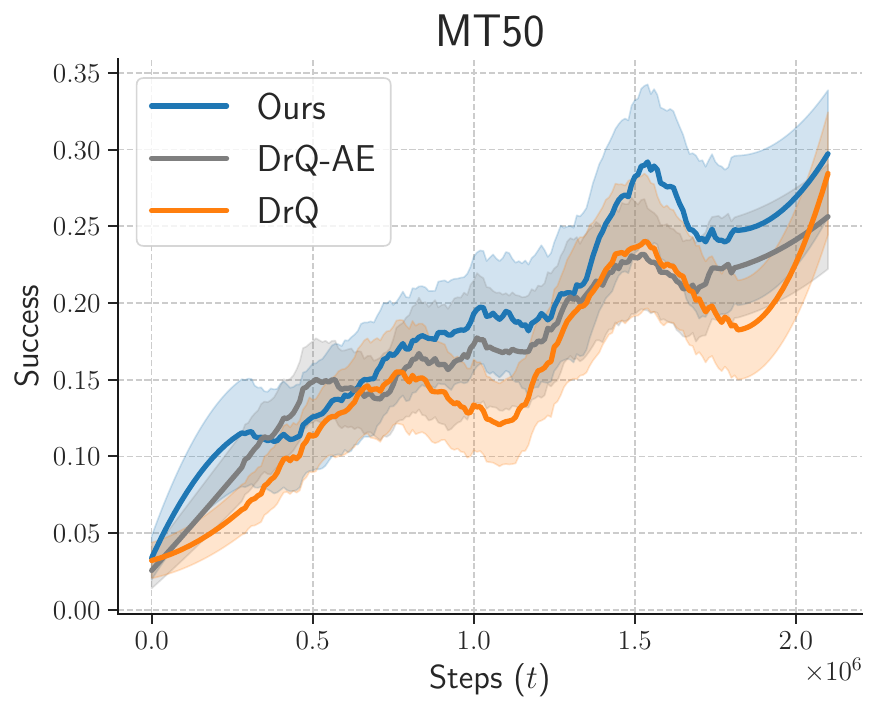}
    \caption{\textbf{Larger-Scale Multi-task setting:} Success for different approaches for joint training on the Meta-World MT50 set of tasks.}
    \label{fig:mt50}
    \vspace{-0.15in}
\end{figure}

\paragraph{Multi-Task Learning} Multi-task environments often use the same agent across multiple tasks. Thus, it is expected that making a distinction between agent and environment features will help multi-task learning as similar agent features can be shared across different tasks. To test this hypothesis, we utilized two small-scale multi-task environment setups: a Meta-World setup with three tasks, and the Franka Kitchen environment with multiple task objects and randomized camera locations (shown in Appendix \ref{appendix:kitchen_cameras}). For each of these, the only input is the raw image. The three Meta-World tasks, button-press-topdown-wall, door-open, and hammer, were selected due to being three visually distinct tasks from our initial set of 10 contact-rich tasks. Results are shown in figure \ref{fig:multi-task}. \ours outperforms baselines in the Meta-World setup and matches performance in the Franka-Kitchen environment. We further evaluate \ours in a larger multi-task setup with the Meta-World MT50 set of tasks, with results shown in figure \ref{fig:mt50}. On this larger set of tasks, \ours matched but did not outperform the baselines. These results are preliminary, and we leave a full investigation into applying and adapting \ours to multi-task environments as future work.  

\section{Discussion and Limitations}

In this paper, we hypothesize that building representations which create a distinction between the agent and environment leads to more effective RL. We do so by formulating a representation learning method that can incorporate agent information directly and efficiently. We construct a learning algorithm, \ours, that encourages representations to learn the agent-environment split by \textit{reconstructing the mask} of the agent. We argue that this form of supervision is a very weak assumption, and in most cases is readily available (even in an approximate form). We show strong results against state of the art approaches in visual RL, on multiple different types of agents (robot arms, hands or even walkers), and in many different and challenging continuous control environments. In our future work, we hope to build agent-centric representations that can allow for building more efficient visual dynamics model, empower exploration and better usage of environment-centric data. Additionally, we would like to investigate disentangled representations for multi-task learning, and real-world robotics (a current limitation of \ours). 

\paragraph{Acknowledgement}
We would like to thank Alexander C. Li and Murtaza Dalal for fruitful discussions. This work is supported by Sony Faculty Research Award and NSF IIS-2024594.

\bibliography{references}

\begin{thebibliography}{54}
\providecommand{\natexlab}[1]{#1}
\providecommand{\url}[1]{\texttt{#1}}
\expandafter\ifx\csname urlstyle\endcsname\relax
  \providecommand{\doi}[1]{doi: #1}\else
  \providecommand{\doi}{doi: \begingroup \urlstyle{rm}\Url}\fi

\bibitem[Arulkumaran et~al.(2017)Arulkumaran, Deisenroth, Brundage, and
  Bharath]{arulkumaran2017deep}
Arulkumaran, K., Deisenroth, M.~P., Brundage, M., and Bharath, A.~A.
\newblock Deep reinforcement learning: A brief survey.
\newblock \emph{IEEE Signal Processing Magazine}, 34\penalty0 (6):\penalty0
  26--38, 2017.

\bibitem[Bahl et~al.(2022)Bahl, Gupta, and Pathak]{bahl2022human2}
Bahl, S., Gupta, A., and Pathak, D.
\newblock Human-to-robot imitation in the wild.
\newblock \emph{RSS}, 2022.

\bibitem[Bahl et~al.(2023)Bahl, Mendonca, Chen, Jain, and
  Pathak]{bahl2023affordances}
Bahl, S., Mendonca, R., Chen, L., Jain, U., and Pathak, D.
\newblock Affordances from human videos as a versatile representation for
  robotics.
\newblock In \emph{CVPR}, 2023.

\bibitem[Chen et~al.(2020)Chen, Kornblith, Norouzi, and
  Hinton]{pmlr-v119-chen20j}
Chen, T., Kornblith, S., Norouzi, M., and Hinton, G.
\newblock A simple framework for contrastive learning of visual
  representations.
\newblock In III, H.~D. and Singh, A. (eds.), \emph{Proceedings of the 37th
  International Conference on Machine Learning}, volume 119 of
  \emph{Proceedings of Machine Learning Research}, pp.\  1597--1607. PMLR,
  13--18 Jul 2020.
\newblock URL \url{https://proceedings.mlr.press/v119/chen20j.html}.

\bibitem[Chen \& He(2021)Chen and He]{chen2021exploring}
Chen, X. and He, K.
\newblock Exploring simple siamese representation learning.
\newblock In \emph{Proceedings of the IEEE/CVF conference on computer vision
  and pattern recognition}, pp.\  15750--15758, 2021.

\bibitem[Dasari \& Gupta(2021)Dasari and Gupta]{dasari2021transformers}
Dasari, S. and Gupta, A.
\newblock Transformers for one-shot visual imitation.
\newblock In \emph{Conference on Robot Learning}, pp.\  2071--2084. PMLR, 2021.

\bibitem[Devin et~al.(2017)Devin, Gupta, Darrell, Abbeel, and
  Levine]{devin2017learning}
Devin, C., Gupta, A., Darrell, T., Abbeel, P., and Levine, S.
\newblock Learning modular neural network policies for multi-task and
  multi-robot transfer.
\newblock In \emph{2017 IEEE international conference on robotics and
  automation (ICRA)}, pp.\  2169--2176. IEEE, 2017.

\bibitem[Duan et~al.(2017)Duan, Andrychowicz, Stadie, Ho, Schneider, Sutskever,
  Abbeel, and Zaremba]{duan2017one}
Duan, Y., Andrychowicz, M., Stadie, B., Ho, O.~J., Schneider, J., Sutskever,
  I., Abbeel, P., and Zaremba, W.
\newblock One-shot imitation learning.
\newblock In \emph{NIPS}, 2017.

\bibitem[Ebert et~al.(2017)Ebert, Finn, Lee, and Levine]{ebert2017self}
Ebert, F., Finn, C., Lee, A.~X., and Levine, S.
\newblock Self-supervised visual planning with temporal skip connections.
\newblock \emph{arXiv preprint arXiv:1710.05268}, 2017.

\bibitem[Finn et~al.(2017)Finn, Yu, Zhang, Abbeel, and Levine]{finn2017one}
Finn, C., Yu, T., Zhang, T., Abbeel, P., and Levine, S.
\newblock One-shot visual imitation learning via meta-learning.
\newblock \emph{CoRL}, 2017.

\bibitem[Fu et~al.(2020)Fu, Kumar, Nachum, Tucker, and Levine]{fu2020d4rl}
Fu, J., Kumar, A., Nachum, O., Tucker, G., and Levine, S.
\newblock D4rl: Datasets for deep data-driven reinforcement learning.
\newblock \emph{arXiv preprint arXiv:2004.07219}, 2020.

\bibitem[Grill et~al.(2020)Grill, Strub, Altch{\'e}, Tallec, Richemond,
  Buchatskaya, Doersch, Avila~Pires, Guo, Gheshlaghi~Azar,
  et~al.]{grill2020bootstrap}
Grill, J.-B., Strub, F., Altch{\'e}, F., Tallec, C., Richemond, P.,
  Buchatskaya, E., Doersch, C., Avila~Pires, B., Guo, Z., Gheshlaghi~Azar, M.,
  et~al.
\newblock Bootstrap your own latent-a new approach to self-supervised learning.
\newblock \emph{Advances in neural information processing systems},
  33:\penalty0 21271--21284, 2020.

\bibitem[Gupta et~al.(2019)Gupta, Kumar, Lynch, Levine, and
  Hausman]{gupta2019relay}
Gupta, A., Kumar, V., Lynch, C., Levine, S., and Hausman, K.
\newblock Relay policy learning: Solving long-horizon tasks via imitation and
  reinforcement learning.
\newblock \emph{arXiv preprint arXiv:1910.11956}, 2019.

\bibitem[Haarnoja et~al.(2018)Haarnoja, Zhou, Abbeel, and
  Levine]{haarnoja2018soft}
Haarnoja, T., Zhou, A., Abbeel, P., and Levine, S.
\newblock Soft actor-critic: Off-policy maximum entropy deep reinforcement
  learning with a stochastic actor.
\newblock \emph{arXiv preprint arXiv:1801.01290}, 2018.

\bibitem[Hafner et~al.(2019)Hafner, Lillicrap, Ba, and
  Norouzi]{hafner2019dream}
Hafner, D., Lillicrap, T., Ba, J., and Norouzi, M.
\newblock Dream to control: Learning behaviors by latent imagination.
\newblock \emph{arXiv preprint arXiv:1912.01603}, 2019.

\bibitem[Hafner et~al.(2020)Hafner, Lillicrap, Norouzi, and
  Ba]{hafner2020mastering}
Hafner, D., Lillicrap, T., Norouzi, M., and Ba, J.
\newblock Mastering atari with discrete world models.
\newblock \emph{arXiv preprint arXiv:2010.02193}, 2020.

\bibitem[He et~al.(2017)He, Gkioxari, Doll{\'a}r, and Girshick]{he2017mask}
He, K., Gkioxari, G., Doll{\'a}r, P., and Girshick, R.
\newblock Mask r-cnn.
\newblock In \emph{ICCV}, 2017.

\bibitem[He et~al.(2020)He, Fan, Wu, Xie, and Girshick]{he2020momentum}
He, K., Fan, H., Wu, Y., Xie, S., and Girshick, R.
\newblock Momentum contrast for unsupervised visual representation learning.
\newblock In \emph{CVPR}, 2020.

\bibitem[He et~al.(2022)He, Chen, Xie, Li, Doll{\'a}r, and
  Girshick]{he2022masked}
He, K., Chen, X., Xie, S., Li, Y., Doll{\'a}r, P., and Girshick, R.
\newblock Masked autoencoders are scalable vision learners.
\newblock In \emph{Proceedings of the IEEE/CVF Conference on Computer Vision
  and Pattern Recognition}, pp.\  16000--16009, 2022.

\bibitem[Hu et~al.(2022)Hu, Huang, Rybkin, and Jayaraman]{hu2022know}
Hu, E.~S., Huang, K., Rybkin, O., and Jayaraman, D.
\newblock Know thyself: Transferable visual control policies through
  robot-awareness.
\newblock In \emph{ICLR 2022 Workshop on Generalizable Policy Learning in
  Physical World}, 2022.

\bibitem[Huang et~al.(2020)Huang, Mordatch, and Pathak]{huang2020one}
Huang, W., Mordatch, I., and Pathak, D.
\newblock One policy to control them all: Shared modular policies for
  agent-agnostic control.
\newblock In \emph{International Conference on Machine Learning}, pp.\
  4455--4464. PMLR, 2020.

\bibitem[Kalashnikov et~al.(2018)Kalashnikov, Irpan, Pastor, Ibarz, Herzog,
  Jang, Quillen, Holly, Kalakrishnan, Vanhoucke, et~al.]{kalashnikov2018qt}
Kalashnikov, D., Irpan, A., Pastor, P., Ibarz, J., Herzog, A., Jang, E.,
  Quillen, D., Holly, E., Kalakrishnan, M., Vanhoucke, V., et~al.
\newblock Qt-opt: Scalable deep reinforcement learning for vision-based robotic
  manipulation.
\newblock \emph{arXiv preprint arXiv:1806.10293}, 2018.

\bibitem[Kalashnikov et~al.(2021)Kalashnikov, Varley, Chebotar, Swanson,
  Jonschkowski, Finn, Levine, and Hausman]{kalashnikov2021mt}
Kalashnikov, D., Varley, J., Chebotar, Y., Swanson, B., Jonschkowski, R., Finn,
  C., Levine, S., and Hausman, K.
\newblock Mt-opt: Continuous multi-task robotic reinforcement learning at
  scale.
\newblock \emph{arXiv preprint arXiv:2104.08212}, 2021.

\bibitem[Kingma \& Welling(2013)Kingma and Welling]{kingma2013auto}
Kingma, D.~P. and Welling, M.
\newblock Auto-encoding variational bayes.
\newblock \emph{arXiv preprint arXiv:1312.6114}, 2013.

\bibitem[Kostrikov et~al.(2020)Kostrikov, Yarats, and
  Fergus]{kostrikov2020image}
Kostrikov, I., Yarats, D., and Fergus, R.
\newblock Image augmentation is all you need: Regularizing deep reinforcement
  learning from pixels.
\newblock \emph{arXiv preprint arXiv:2004.13649}, 2020.

\bibitem[Laskin et~al.(2020{\natexlab{a}})Laskin, Lee, Stooke, Pinto, Abbeel,
  and Srinivas]{laskin2020reinforcement}
Laskin, M., Lee, K., Stooke, A., Pinto, L., Abbeel, P., and Srinivas, A.
\newblock Reinforcement learning with augmented data.
\newblock \emph{Advances in neural information processing systems},
  33:\penalty0 19884--19895, 2020{\natexlab{a}}.

\bibitem[Laskin et~al.(2020{\natexlab{b}})Laskin, Srinivas, and
  Abbeel]{pmlr-v119-laskin20a}
Laskin, M., Srinivas, A., and Abbeel, P.
\newblock {CURL}: Contrastive unsupervised representations for reinforcement
  learning.
\newblock In III, H.~D. and Singh, A. (eds.), \emph{Proceedings of the 37th
  International Conference on Machine Learning}, volume 119 of
  \emph{Proceedings of Machine Learning Research}, pp.\  5639--5650. PMLR,
  13--18 Jul 2020{\natexlab{b}}.
\newblock URL \url{https://proceedings.mlr.press/v119/laskin20a.html}.

\bibitem[Levine et~al.(2016)Levine, Finn, Darrell, and Abbeel]{levineFDA15}
Levine, S., Finn, C., Darrell, T., and Abbeel, P.
\newblock End-to-end training of deep visuomotor policies.
\newblock \emph{JMLR}, 2016.

\bibitem[Lillicrap et~al.(2015)Lillicrap, Hunt, Pritzel, Heess, Erez, Tassa,
  Silver, and Wierstra]{lillicrap2015continuous}
Lillicrap, T.~P., Hunt, J.~J., Pritzel, A., Heess, N., Erez, T., Tassa, Y.,
  Silver, D., and Wierstra, D.
\newblock Continuous control with deep reinforcement learning.
\newblock \emph{arXiv preprint arXiv:1509.02971}, 2015.

\bibitem[Mendonca et~al.(2023)Mendonca, Bahl, and Pathak]{mendonca2023alan}
Mendonca, R., Bahl, S., and Pathak, D.
\newblock Alan: Autonomously exploring robotic agents in the real world.
\newblock \emph{arXiv preprint arXiv:2302.06604}, 2023.

\bibitem[Nair et~al.(2018)Nair, Pong, Dalal, Bahl, Lin, and
  Levine]{nair2018visual}
Nair, A.~V., Pong, V., Dalal, M., Bahl, S., Lin, S., and Levine, S.
\newblock Visual reinforcement learning with imagined goals.
\newblock In \emph{NeurIPS}, pp.\  9191--9200, 2018.

\bibitem[Nair et~al.(2022)Nair, Rajeswaran, Kumar, Finn, and Gupta]{r3m}
Nair, S., Rajeswaran, A., Kumar, V., Finn, C., and Gupta, A.
\newblock R3m: A universal visual representation for robot manipulation.
\newblock \emph{arXiv preprint arXiv:2203.12601}, 2022.

\bibitem[Neumann et~al.(2014)Neumann, Daniel, Paraschos, Kupcsik, and
  Peters]{neumann2014learning}
Neumann, G., Daniel, C., Paraschos, A., Kupcsik, A., and Peters, J.
\newblock Learning modular policies for robotics.
\newblock \emph{Frontiers in computational neuroscience}, 8:\penalty0 62, 2014.

\bibitem[Parisi et~al.(2021)Parisi, Dean, Pathak, and
  Gupta]{parisi2021interesting}
Parisi, S., Dean, V., Pathak, D., and Gupta, A.
\newblock Interesting object, curious agent: Learning task-agnostic
  exploration.
\newblock \emph{Advances in Neural Information Processing Systems},
  34:\penalty0 20516--20530, 2021.

\bibitem[Pathak et~al.(2016)Pathak, Krahenbuhl, Donahue, Darrell, and
  Efros]{pathak2016context}
Pathak, D., Krahenbuhl, P., Donahue, J., Darrell, T., and Efros, A.~A.
\newblock Context encoders: Feature learning by inpainting.
\newblock In \emph{CVPR}, 2016.

\bibitem[Peters et~al.(2010)Peters, M\"{u}lling, and Alt\"{u}n]{peters2010reps}
Peters, J., M\"{u}lling, K., and Alt\"{u}n, Y.
\newblock Relative entropy policy search.
\newblock In \emph{Proceedings of the Twenty-Fourth AAAI Conference on
  Artificial Intelligence}, AAAI’10, pp.\  1607–1612. AAAI Press, 2010.

\bibitem[Pong et~al.(2019)Pong, Dalal, Lin, Nair, Bahl, and
  Levine]{pong2019skew}
Pong, V.~H., Dalal, M., Lin, S., Nair, A., Bahl, S., and Levine, S.
\newblock Skew-fit: State-covering self-supervised reinforcement learning.
\newblock \emph{arXiv preprint arXiv:1903.03698}, 2019.

\bibitem[Radosavovic et~al.(2022)Radosavovic, Xiao, James, Abbeel, Malik, and
  Darrell]{radosavovic2022real}
Radosavovic, I., Xiao, T., James, S., Abbeel, P., Malik, J., and Darrell, T.
\newblock Real-world robot learning with masked visual pre-training.
\newblock \emph{arXiv preprint arXiv:2210.03109}, 2022.

\bibitem[Rajeswaran et~al.(2017)Rajeswaran, Kumar, Gupta, Vezzani, Schulman,
  Todorov, and Levine]{rajeswaran2017learning}
Rajeswaran, A., Kumar, V., Gupta, A., Vezzani, G., Schulman, J., Todorov, E.,
  and Levine, S.
\newblock Learning complex dexterous manipulation with deep reinforcement
  learning and demonstrations.
\newblock \emph{arXiv preprint arXiv:1709.10087}, 2017.

\bibitem[Rezende et~al.(2014)Rezende, Mohamed, and
  Wierstra]{rezende2014stochastic}
Rezende, D.~J., Mohamed, S., and Wierstra, D.
\newblock Stochastic backpropagation and approximate inference in deep
  generative models.
\newblock In \emph{International conference on machine learning}, pp.\
  1278--1286. PMLR, 2014.

\bibitem[Shapiro(2010)]{shapiro2010embodied}
Shapiro, L.
\newblock \emph{Embodied cognition}.
\newblock Routledge, 2010.

\bibitem[Shaw et~al.(2022)Shaw, Bahl, and Pathak]{shaw2022video}
Shaw, K., Bahl, S., and Pathak, D.
\newblock Videodex: Learning dexterity from internet videos.
\newblock In \emph{CoRL}, 2022.

\bibitem[Stone et~al.(2021)Stone, Ramirez, Konolige, and
  Jonschkowski]{stone2021distracting}
Stone, A., Ramirez, O., Konolige, K., and Jonschkowski, R.
\newblock The distracting control suite--a challenging benchmark for
  reinforcement learning from pixels.
\newblock \emph{arXiv preprint arXiv:2101.02722}, 2021.

\bibitem[Tassa et~al.(2018)Tassa, Doron, Muldal, Erez, Li, Casas, Budden,
  Abdolmaleki, Merel, Lefrancq, et~al.]{tassa2018deepmind}
Tassa, Y., Doron, Y., Muldal, A., Erez, T., Li, Y., Casas, D. d.~L., Budden,
  D., Abdolmaleki, A., Merel, J., Lefrancq, A., et~al.
\newblock Deepmind control suite.
\newblock \emph{arXiv preprint arXiv:1801.00690}, 2018.

\bibitem[Todorov et~al.(2012)Todorov, Erez, and Tassa]{todorov12mujoco}
Todorov, E., Erez, T., and Tassa, Y.
\newblock {MuJoCo: A physics engine for model-based control}.
\newblock In \emph{IROS}, 2012.

\bibitem[Wang et~al.(2018)Wang, Liao, Ba, and Fidler]{wang2018nervenet}
Wang, T., Liao, R., Ba, J., and Fidler, S.
\newblock Nervenet: Learning structured policy with graph neural networks.
\newblock 2018.

\bibitem[Watter et~al.(2015)Watter, Springenberg, Boedecker, and
  Riedmiller]{watter2015embed}
Watter, M., Springenberg, J., Boedecker, J., and Riedmiller, M.
\newblock Embed to control: A locally linear latent dynamics model for control
  from raw images.
\newblock In \emph{NIPS}, 2015.

\bibitem[Xiao et~al.(2022)Xiao, Radosavovic, Darrell, and Malik]{mvp}
Xiao, T., Radosavovic, I., Darrell, T., and Malik, J.
\newblock Masked visual pre-training for motor control.
\newblock \emph{arXiv preprint arXiv:2203.06173}, 2022.

\bibitem[Yarats et~al.(2021{\natexlab{a}})Yarats, Fergus, Lazaric, and
  Pinto]{yarats2021drqv2}
Yarats, D., Fergus, R., Lazaric, A., and Pinto, L.
\newblock Mastering visual continuous control: Improved data-augmented
  reinforcement learning.
\newblock \emph{arXiv preprint arXiv:2107.09645}, 2021{\natexlab{a}}.

\bibitem[Yarats et~al.(2021{\natexlab{b}})Yarats, Zhang, Kostrikov, Amos,
  Pineau, and Fergus]{yarats2021improving}
Yarats, D., Zhang, A., Kostrikov, I., Amos, B., Pineau, J., and Fergus, R.
\newblock Improving sample efficiency in model-free reinforcement learning from
  images.
\newblock In \emph{Proceedings of the AAAI Conference on Artificial
  Intelligence}, volume~35, pp.\  10674--10681, 2021{\natexlab{b}}.

\bibitem[Yu et~al.(2018)Yu, Finn, Xie, Dasari, Zhang, Abbeel, and
  Levine]{yu2018one}
Yu, T., Finn, C., Xie, A., Dasari, S., Zhang, T., Abbeel, P., and Levine, S.
\newblock One-shot imitation from observing humans via domain-adaptive
  meta-learning.
\newblock \emph{RSS}, 2018.

\bibitem[Yu et~al.(2020)Yu, Quillen, He, Julian, Hausman, Finn, and
  Levine]{yu2020meta}
Yu, T., Quillen, D., He, Z., Julian, R., Hausman, K., Finn, C., and Levine, S.
\newblock Meta-world: A benchmark and evaluation for multi-task and meta
  reinforcement learning.
\newblock In \emph{Conference on robot learning}, pp.\  1094--1100. PMLR, 2020.

\bibitem[Zhang et~al.(2019)Zhang, Vikram, Smith, Abbeel, Johnson, and
  Levine]{zhang2019solar}
Zhang, M., Vikram, S., Smith, L., Abbeel, P., Johnson, M., and Levine, S.
\newblock Solar: Deep structured representations for model-based reinforcement
  learning.
\newblock In \emph{International conference on machine learning}, pp.\
  7444--7453. PMLR, 2019.

\bibitem[Zhu et~al.(2020)Zhu, Yu, Gupta, Shah, Hartikainen, Singh, Kumar, and
  Levine]{zhu2020ingredients}
Zhu, H., Yu, J., Gupta, A., Shah, D., Hartikainen, K., Singh, A., Kumar, V.,
  and Levine, S.
\newblock The ingredients of real-world robotic reinforcement learning.
\newblock \emph{arXiv preprint arXiv:2004.12570}, 2020.

\end{thebibliography}
\bibliographystyle{icml2023}

\newpage
\appendix
\onecolumn

\section{Theoretical Justification}

\subsection{Representation Objective}
\label{appendix:rep_obj}

We describe how we obtain the training objective for the representation space from the graphical model shown in Figure \ref{fig:pgm}. In order to learn $z,z_R$, we maximize $J = \log p(X, X_R)$, and learn a variational approximation $q_\theta(z, z_R|X)$ to the posterior. 
We have : 
\begin{flalign*}
    \log p(X, X_R) & = \log \int_{z, z_R} p(z). p(z_R|z). p(X|z). p(X_R|z_R) &\\
    & \geq \int_{z, z_R} \log( p (z, z_R). p (X|z). p(X_R|z_R))&\\
    & =\mathop{\mathbb{E}}_{z,z_R \sim q} \left[ \log \left( \frac{p(z, z_R).p(X|z).p(X_R|z_R)}{q(z,z_R|X)} \right) \right]&\\
    & =\mathop{\mathbb{E}}_{z,z_R \sim q} \left[ \log p(X|z) \right] + \mathop{\mathbb{E}}_{z,z_R \sim q} \left[ \log p(X_R|z_R) \right] - D_{KL}(q(z,z_R|X))||(p(z,z_R))
\end{flalign*}

\subsection{RL Performance}
\label{appendix:rl_performance}

Consider the general value function described in Section 4: 

\begin{equation}
    V(\pmb{z}_t) = V_R(\pmb{z}^R_t) +  V_C(\pmb{z_t})
\end{equation}

Here $z_t$ is the concatenation of $\pmb{z}^R_t$ and $\pmb{z}^E_t$. We can think of the following

\begin{flalign*}
    \pmb{z}^R_t \sim \mathcal{N}(\mu_R, \sigma_R) &\\
    \pmb{z}^E_t \sim \mathcal{N}(\mu_E, \sigma_E) &\\
\end{flalign*}

Let us assume two cases, one where there is environment interaction and one where there is none. In the case where there is very little or no interaction, we can think in this case that $\sigma_R$ will be small, as there is less noise in the robot information, as these features are more salient for the value function. Assuming that the value function can be written as a quadratic function: 

\begin{flalign*}
V(\pmb{z}_t) = [\pmb{z}^R_t, \pmb{z}^E_t]^TV[\pmb{z}^R_t, \pmb{z}^E_t] & \\ 
V(\pmb{z}_t) = [\pmb{z}^R_t, \pmb{z}^E_t]^T[V_1, V_2][\pmb{z}^R_t, \pmb{z}^E_t] & \\ 
V(\pmb{z}_t) = (\pmb{z}^{R}_t)^T V_1\pmb{z}^R_t + (\pmb{z}^{E}_t)^TV_1\pmb{z}^E_t + 2(\pmb{z}^{R}_t)^TV_2\pmb{z}^{E}_t
\end{flalign*}

When fitting these parameters, due to the high noise in $\pmb{z}^E_t$, we find that singular values of $V_1$ are much higher than those of $V_2$. 

Now let us look at the observation $X$. This can be thought of as: 

\begin{equation}
\label{eq:obs_proj}
X = Q\pmb{z}_t + \epsilon
\end{equation}

Where $Q$ is some matrix in $SO(3)$ and $\epsilon \sim \mathcal{N}(0, 1)$. When learning $\pmb{z}_t$ from these, in the case of an \textit{autoencoder}, we will fit $\hat{\pmb{z}}$ to the top principal components of $\pmb{z}^E_t$, as $\sigma_E > \sigma_R$. However, in the case where we are explicitly reconstructing $X_R$, the model will fit $\hat{\pmb{z}}$ to a mix of top principal components of $\pmb{z}^R_t$ and $\pmb{z}^E_t$, which means it is better at recovering the true $\pmb{z}_t$. 

\subsection{Toy Experiment}
\label{appendix:toy_experiment}

To showcase the intuition that it can be easier to learn the value function/policy if the latent representation directly decouples the agent and environment state, we construct the following toy experiment. We create a 2D continuous point mass environment with a changing obstacle position and a randomly sampled goal and initialization position. The ground truth reward is made up of an attraction term (exponential distance to goal) and a repulsion term (exponential distance to obstacle). The agent does not observe the ground truth state, but a high dimensional projection of a noisy latent state as mentioned in equation \ref{eq:obs_proj}. Furthermore, as in Appendix \ref{appendix:rl_performance}, $\pmb{z}^R_t$ and $\pmb{z}^E_t$ are drawn from normal distributions, where $\mu_R$ is the position of the agent and $\mu_E$ contains the position of the goal and the position and size of the obstacle.

When running SEAR, the agent additionally has access to the mask : $X_R = Q_2 z_R.$

We collect data in this environment to learn $\hat{z}$ from $X$ and $\hat{z}_R$ from $X_R$, which learn the overall structure of the data and the agent-relevant structure respectively. We use this to train a predictive reward model: $V(\hat{z}, \hat{z}_R)$, which is later used for MPC. We test the performance of SEAR which models ($\hat{z}, \hat{z}_R$), to a baseline which directly models $\hat{z}$, using final distance to goal as the metric, for different relative values of $\sigma_R$ and $\sigma_E$, presenting the results in Table \ref{table:toy_results}.

\begin{table}[h!] \centering
    \caption{Final distance to goal for varying relative values of agent and environment observation noise}
    \begin{tabular}{ |l|c|c| } 
        \hline
        & $V(\hat{z})$ & $V(\hat{z}, \hat{z}_R)$ \\
        \hline
        $\sigma_E > \sigma_R$ & 1.35 & 1.09 \\
        $\sigma_E = \sigma_R$ & 1.09 & 1.08 \\
        $\sigma_E < \sigma_R$ & 1.40 & 1.36 \\
        \hline
    \end{tabular}
    \label{table:toy_results}
\end{table}

Table \ref{table:toy_results} shows the distance to goal achieved for different noise levels by using $V(\hat{z})$ or $V(\hat{z}, \hat{z}_R)$ for planning. We see that when the noise in the environment observation is high, SEAR provides much more effective control, and that the performance is roughly equal in all other cases.

\section{Limitations and Broader Impacts}

\ours has a couple of different limitations. First, it's performance on learning representations for multi-task learning is not fully clear. We showed preliminary results on two small-scale multi-task settings and the Meta-World MT50 setting, but further investigation is required. \ours has not yet been tested on real-world environments. 

\ours was only tested with the DrQv2 algorithm, and so while \ours can be deployed with any reinforcement learning algorithm with an image encoder, future work still needs to explore the performance of using \ours with other algorithms. 

\ours does not provide any benefit over other approaches in cases where the robot is not visible in the input image. Thus, future work will need to investigate how to incorporate such visual observations, which may be common in cases where, for example, the robot has a first-person view.

Another limitation of \ours is that the representations learned by \ours may not be closely aligned with human interests. This becomes more relevant as visual control algorithms are deployed in greater numbers to perform a wider variety of tasks, especially for tasks with a direct impact on humans. 

Even if the representations learned are relatively well-aligned with the intentions of a designer using this algorithm, there are no safe-guards against a designer purposefully using \ours to learn visual control tasks that can cause harm to humans. 

\ours could also be used to aid the deployment of more general-purpose robots, which may affect people's employment and overall economic conditions.

\section{Implementation Details}

Here, we present additional details relevant to our implementation of \ours. 

\subsection{Code}

The code of ours can be found at \url{https://sear-rl.github.io}

\subsection{Hyperparameters}

For \ours, we used the DrQv2 algorithm
\cite{yarats2021drqv2} with a modified encoder, and added additional mask and reconstruction decoders. A list of hyperparameters can be found in Table \ref{table:hyperparams}. Most hyperparameters have the same value as in \cite{yarats2021drqv2}. 

Compared to the original replay buffer size of 1e6, we used a smaller replay buffer size of 2.5e5. This change was made mainly to reduce RAM usage for each experiment. For instance, a buffer of size 1e6 with a frame stack of 3 corresponds to 3e6 rgb images of size (84, 84). Given 3 bytes per pixel, this gives us a total RAM usage of about 63.5 GB. This problem could be alleviated through tricks such as only storing each image once, instead of duplicating the image 3 times for a frame stack of 3. However, we did not see any performance decrease during preliminary experiments when reducing the replay buffer. We used this smaller replay buffer size as well when evaluating baselines. 

Another difference in value between hyperparameters common to both DrQv2 and \ours is action repeat. On the Franka Kitchen and Hand Manipulation Suite environments, we set the action repeat to 1. We did not experiment with varying the action repeat, but we did use the same action repeat in our comparison baselines. 

\ours has a couple of extra hyperparameters that are not present in DrQv2. One such hyperparameter is the learning rate for the two decoders. For simplicity, we ended up using the same learning rate as the actor and critic. The other new hyperparameters are the coefficients controlling the contribution of the reconstruction and mask decoders to the total loss, relative to the critic loss. These correspond to coefficients $c_1$ and $c_2$ in equation \ref{eq:total_loss}. We observed that the cross entropy mask loss tended to converge to a value about four times greater than the reconstruction loss, so we set the reconstruction loss scaling coefficient to be four times the mask loss scaling coefficient. This kept the contribution of the decoders relatively equal. From the ablation shown in \ref{fig:abl-misc}, \ours is relatively insensitive to the choice for $c_1$. 

\begin{table} 
    \centering
    \caption{Hyperparameter Settings}
    \begin{tabular}{ |l|c| } 
        \hline
        Hyperparameter & Value \\
        \hline
        Replay Buffer Size & 2.5e5 \\
        Frame Stack & 3 \\ 
        Action Repeat & 2 - Meta-World; Distracting Control \\
        & 1 - Kitchen; Hand Manipulation\\
        Seed Frames & 4000 \\
        Exploration Steps & 2000 \\
        N-Step Return & 3 \\
        Batch Size & 256 \\
        Discount & 0.99 \\
        Optimizer & Adam \\
        Actor, Critic Learning Rate & 1e-4\\
        Decoder Learning Rate & 1e-4 \\
        Agent Update Frequency & 2 \\
        Critic Q-function Soft-Update Rate $\tau$ & 0.01 \\
        Actor, Critic Feature Dimensions & 50 \\
        Latent Dimensions & 4096 - Single-task \\
        & 512 - Multi-task, Transfer \\
        Exploration StdDev. Schedule & linear(1.0,0.1,2e6) - Meta-World MT \\
        & linear(1.0,0.1,5e5) - otherwise\\
        Exploration StdDev Clip & 0.3 \\
        Observation Render Size & (84, 84)\\
        Reconstruction Loss Scaling Coefficient ($c_1$) & 0.01 \\
        Mask Loss Scaling Coefficient ($c_2$) & 0.0025 \\
        Camera Name/ID & "Corner" - Meta-World \\
        & "Fixed" - Hand Manipulation; Kitchen\\
        & 0 - Distracting Control\\
        Evaluation Frequency & 1e4 \\
        Evaluation Episodes & 10 \\
        \hline
    \end{tabular}
    \label{table:hyperparams}
\end{table}

\subsection{Architecture}

Our architecture differs from \cite{yarats2021drqv2} with the addition of two decoders, and a modification to the encoder. The architectures used for the actor and critic are the same.

Similar to \cite{yarats2021drqv2}, our encoder architecture contains a series of four convolutions with ReLU's between them. Each convolution has a kernel size of 3, and every convolution except the first one has a stride of 1. The first convolution has a stride of 2. Every convolution has 32 feature maps. This transforms a stack of 84x84 rgb images into a 32x35x35 set of activation maps. After the convolutions, we add a 2D average pooling layer with kernel size of 4 and stride of 4. This helps reduce the overall dimensionality down, giving us 32x8x8 features. We flatten this set of feature maps, and pass this through a linear layer, projecting the activations to a desired latent dimension. The encoder outputs the output from this linear layer.

The decoder architecture, used for both the mask and reconstruction decoders, starts with a linear layer with ReLU activation. This maps the latent vectors to a 2048 dimensional vector, which we reshape into a 32x8x8 set of feature maps. We pass these feature maps through 5 transpose convolutions. The first transpose convolution in the decoder is set up to mirror the average pooling layer in the encoder. It has a kernel size and stride of 4. We used 16 channels for this convolution, as our original encoder implementation had no linear projection, and so each half of the latent vector passed to each decoder only came from 16 of the 32 channels output by the average pooling layer. After adding a linear projection to the encoder, each half of the latent now comes from a combination of all 32 channels output by the average pooling layer, but we kept using 16 channels in the first layer of our decoder. Future work could change the first transpose convolution to use 32 channels, although we do not expect such a change to make any significant impact on the overall performance of \ours. The last four transpose convolutions mirror the 4 convolutions in the encoder. They have 32 channels, except for the last convolution, which has the same number of channels as the stack of rgb images for the reconstruction decoder, or the same number of channels as the stack of masks for the mask decoder. They have a kernel size of 3, and a stride of 1, except for the last transpose convolution which has a stride of 2. The output from the transpose convolutions is passed through an output activation, which is the identity for the reconstruction decoder and a sigmoid for the mask decoder. 

\subsection{Training Details}

We train each model using a mix of RTX6000, A5000, A6000 or 2080Ti GPUs. Each run takes about 3-7 hours of training on these. 

\subsection{Robot Mask Preprocessing} \label{appendix:mask_preprocessing}
For segmentation masks, single pixel wide artifacts were observed to render around the edges of some non-robot objects. At a resolution of 84x84, some robot features were also on the order of a single pixel wide, which made filtering at such a resolution more difficult. We observed that the artifacts remained one-pixel wide, even as the resolution was increased. Thus, to filter out these artifacts, segmentation masks were rendered at a resolution of (252, 252), a morphological opening was applied, and then the masks were downsampled to a resolution of (84, 84) before being passed to the agent.  

\subsection{Noisy and Approximate Mask Generation} \label{appendix:noisy_mask_generation}

Noisy masks were generated by taking the clean, filtered robot mask, and randomly setting robot labels to non-robot labels with a user-specified probability. To generate an approximate mask, we take the original robot mask, downsample it, and then upsample it back to its original size. We then apply gaussian blurring and threshold the image to get a new mask.

\section{Real-World Robot Segmentation Model}
\label{appendix:real_robot_masks}

\begin{figure*}[h!]
    \centering
    \includegraphics[width=0.9\linewidth]{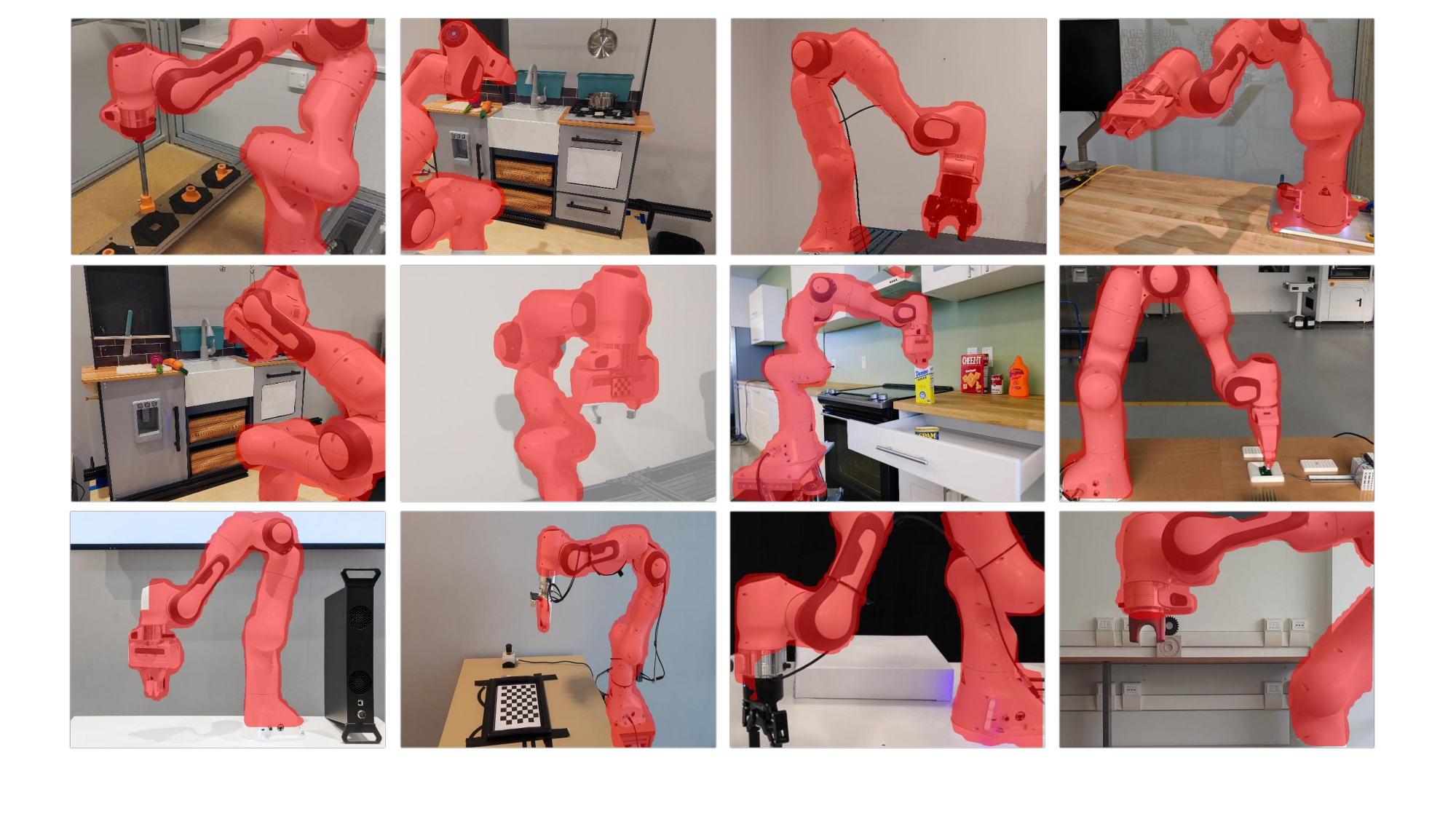}
    \caption{Segmentation masks predicted on out-of-distribution images by a fine-tuned Mask-RCNN model \cite{he2017mask}. } 
    \label{fig:real_robot_masks}
\end{figure*}

In order to more directly test applicability to real robots, we train a segmentation model for a real Franka Panda robot. For this setup, we finetune a Mask-RCNN model \cite{he2017mask} on around 100 images of our own robot (Franka Panda) as well as a few internet images, both of which we manually label. Figure \ref{fig:real_robot_masks} shows images of the masks obtained by running our model on out-of-distribution robot images as well as new viewpoints. When paired with the results from \ref{subsection:AnalysisAndAblations}, which show that \ours can work with noisy and approximate robot masks, we have hope that \ours may also be applicable to training in real-world settings. 

\clearpage

\section{Decoder Ablations}
\label{appendix:decoder_ablations}

\begin{figure}[h!]
    \centering
    \includegraphics[width=\linewidth]{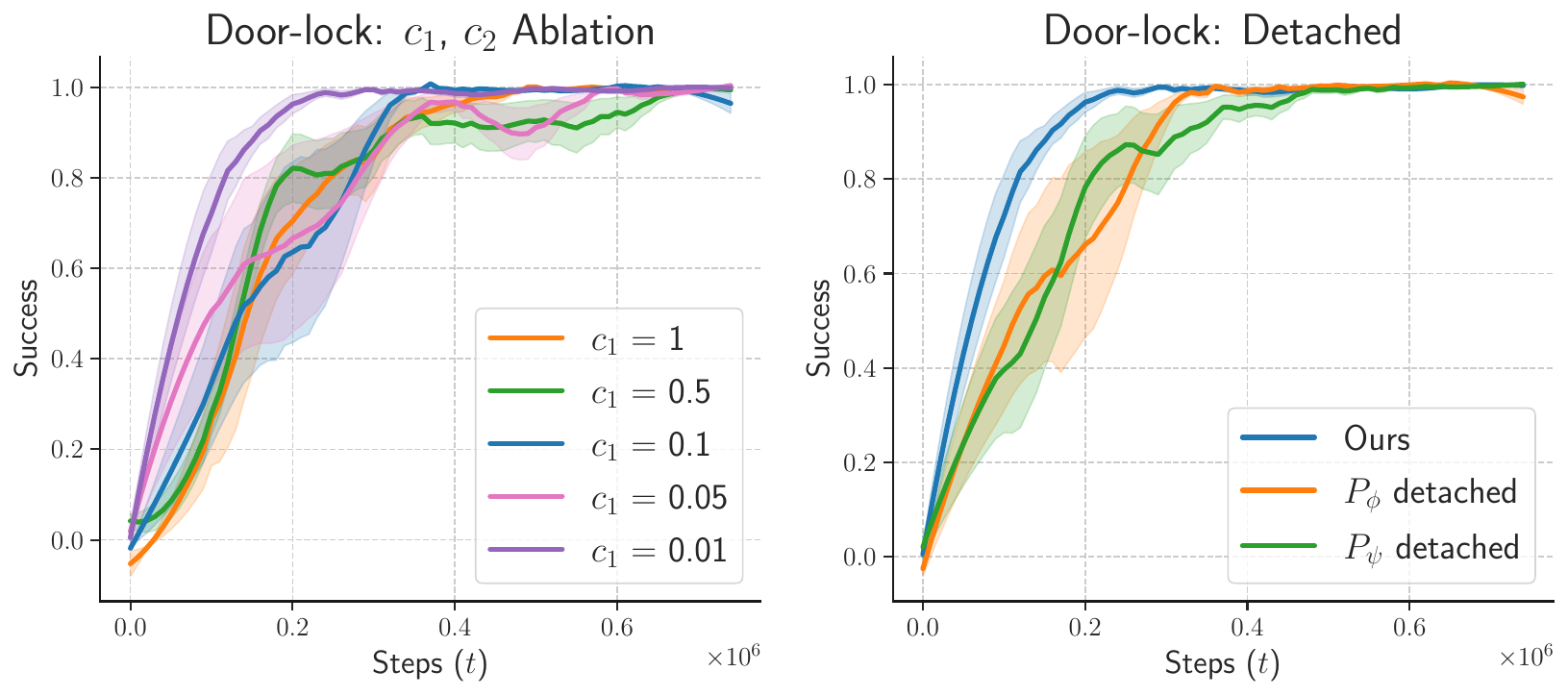}
    \caption{Ablation of hyperparameters of \ours such as the coefficients of the reconstruction loss: $c_1$, $c_2$.}
    \label{fig:abl-misc}
    \vspace{-0.1in}
\end{figure}

\section{Auxillary Mask Loss Without Latent Separation}
\label{appendix:no_zr}

\begin{figure}[h!]
    \centering
    \includegraphics[width=\linewidth]{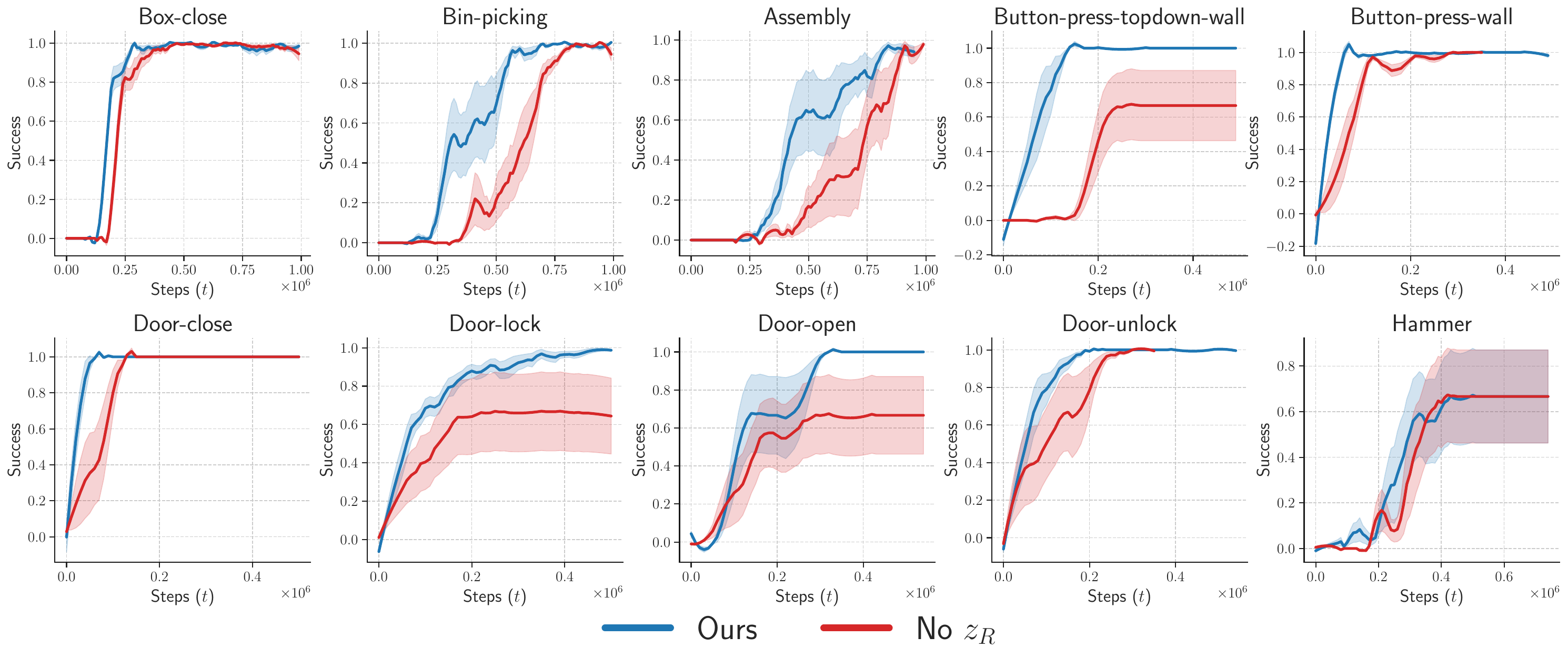}
    \caption{Ablation comparing \ours to a model where you add an auxillary mask loss but do not split the latent vector.}
    \label{fig:abl-misc}
    \vspace{-0.1in}
\end{figure}

\clearpage
\section{Encoder Activation Maps}
\label{appendix:activation_maps}

\begin{figure*}[h!]
    \centering
    \includegraphics[width=0.9\linewidth]{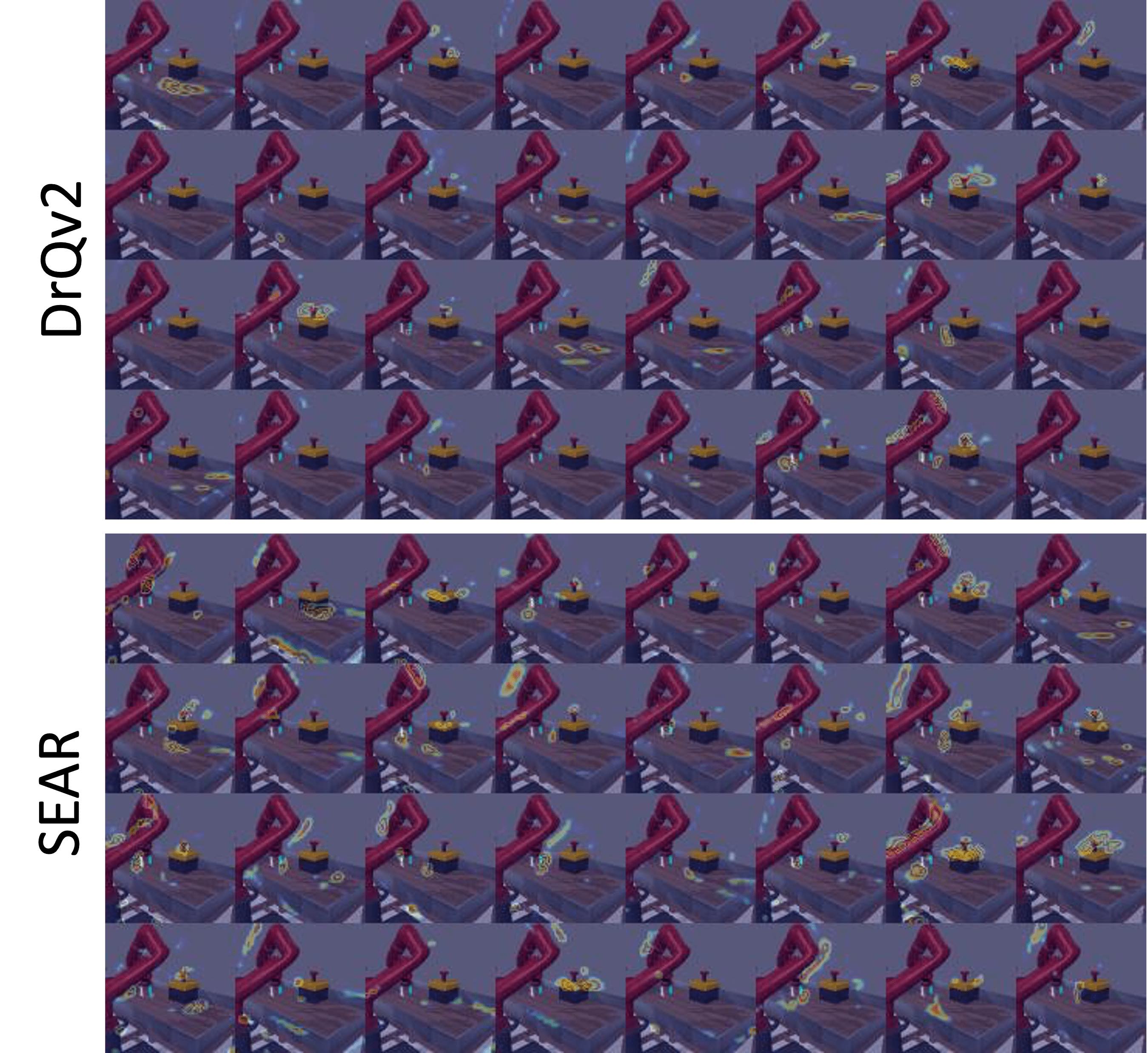}
    \caption{Activation maps from the fourth convolution of the encoder, resized and overlaid on top of the corresponding input image.} 
    \label{fig:mt_button_act_full}
    \vspace{-0.1in}
\end{figure*}

\begin{figure*}[h!]
    \centering
    \includegraphics[width=0.9\linewidth]{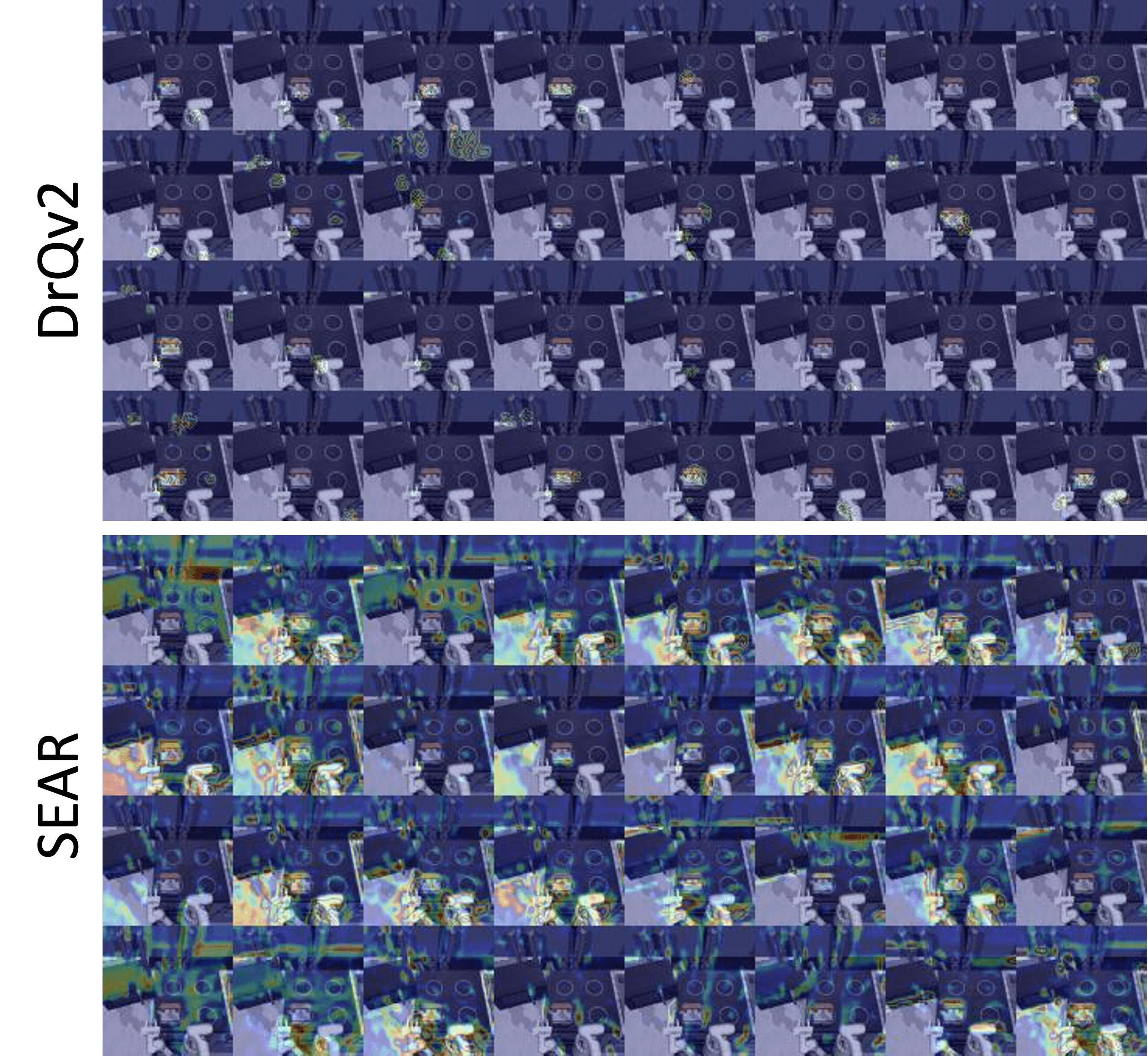}
    \caption{Activation maps from the fourth convolution of the encoder, resized and overlaid on top of the corresponding input image.} 
    \label{fig:kitchen_act_full}
\end{figure*}

\begin{figure*}[h!]
    \centering
    \includegraphics[width=0.9\linewidth]{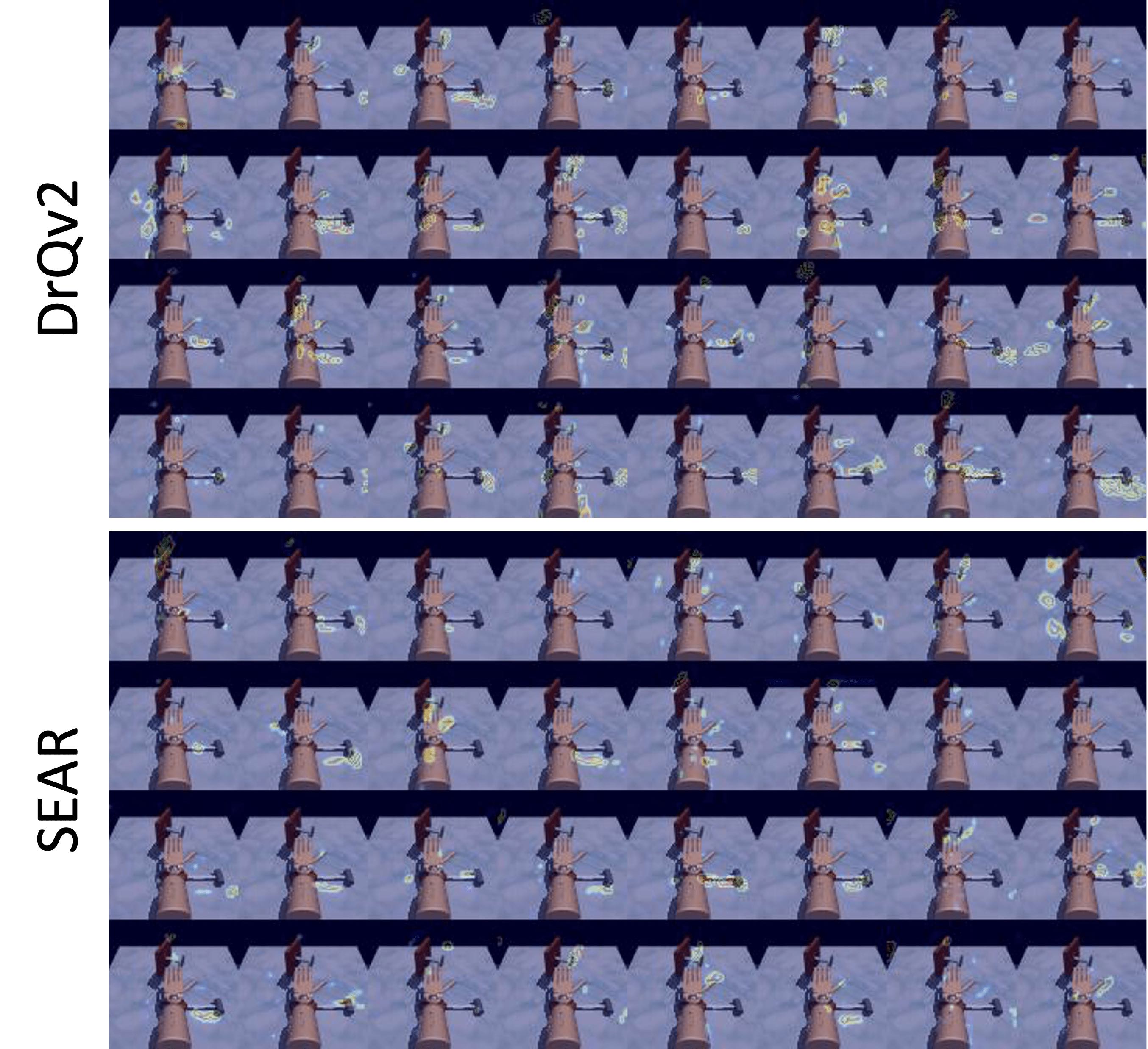}
    \caption{Activation maps from the fourth convolution of the encoder, resized and overlaid on top of the corresponding input image.} 
    \label{fig:adroit_act_full}
\end{figure*}

\clearpage
\section{Multi-Task Franka Kitchen Camera Locations}
\label{appendix:kitchen_cameras}

\begin{figure*}[h!]
    \centering
    \includegraphics[width=0.9\linewidth]{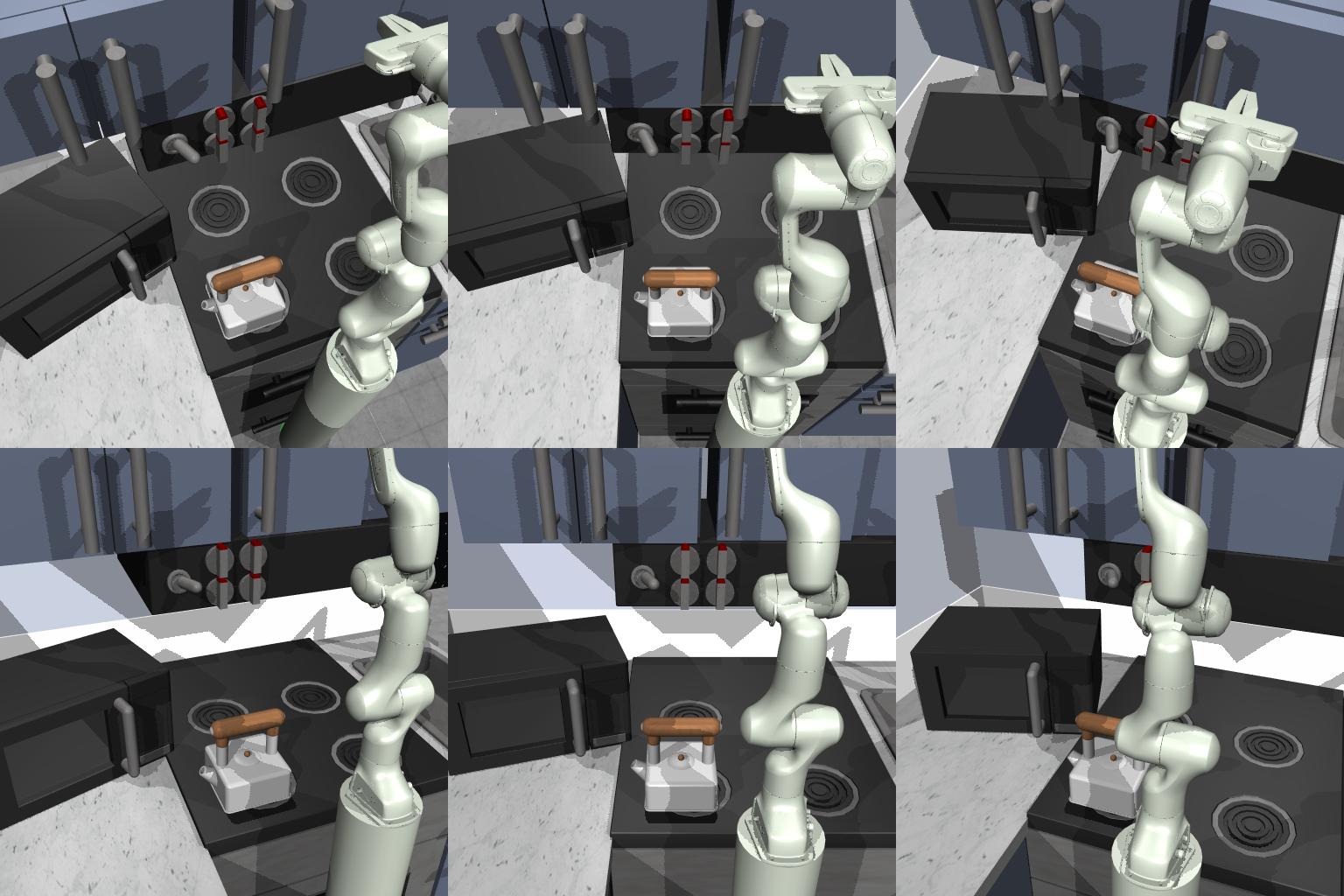}
    \caption{Set of 6 camera angles randomly selected from each episode in the Franka Kitchen multi-task setup} 
    \label{fig:kitchen_all_cams}
\end{figure*}

\end{document}